\documentclass[sigconf]{acmart}

\usepackage{booktabs} 

\usepackage{balance}
\usepackage{url,graphicx}
\usepackage{mathrsfs, amsfonts}
\usepackage{multirow}

\usepackage{enumitem}
\setlist[itemize]{noitemsep, topsep=0pt}

\usepackage{balance}
\usepackage{graphicx}
\usepackage{caption}
\usepackage{subcaption}

\usepackage{algorithm}
\usepackage{algorithmic}

\usepackage[flushleft]{threeparttable}

\usepackage{comment}
\newcommand{\bing}[1]{\textbf{[BING: #1]}}

\copyrightyear{2019}
\acmYear{2019} 
\setcopyright{iw3c2w3}
\acmConference[WWW '19]{Proceedings of the 2019 World Wide Web Conference}{May 13--17, 2019}{San Francisco, CA, USA}
\acmBooktitle{Proceedings of the 2019 World Wide Web Conference (WWW '19), May 13--17, 2019, San Francisco, CA, USA}
\acmPrice{}
\acmDOI{10.1145/3308558.3313496}
\acmISBN{978-1-4503-6674-8/19/05}

\begin{document}

	\title{Persona-Aware Tips Generation}
	\titlenote{The work described in this paper was partially supported by grants from the Research Grant Council of the Hong Kong Special Administrative Region, China (Project Codes: 14203414) and the Direct Grant of the Faculty of Engineering, CUHK (Project Code: 4055093).}
	
	\author{Piji Li}
	\affiliation{%
		\institution{$^1$Tencent AI Lab}
		\city{Shenzhen}
		\country{China}\\
		\institution{$^2$The Chinese University of Hong Kong}
		\city{Hong Kong}
		\country{China}
	}
	\email{pijili@tencent.com}
	
	\author{Zihao Wang}
	\affiliation{%
		\institution{The Chinese University of Hong Kong}
		\city{Hong Kong}
		\country{China}
	}
	\email{zhwang@se.cuhk.edu.hk}
	
	\author{Lidong Bing}
	\affiliation{
		\institution{R\&D Center Singapore }
		\country{Alibaba DAMO Academy}}
	\email{l.bing@alibaba-inc.com}
	
	\author{Wai Lam}
	\affiliation{%
		\institution{The Chinese University of Hong Kong}
		\city{Hong Kong}
		\country{China}
	}
	\email{wlam@se.cuhk.edu.hk}

\begin{abstract}
Tips, as a compacted and concise form of reviews, were paid less attention by researchers. In this paper, we investigate the task of tips generation by considering the ``persona'' information which captures the intrinsic language style of the users or the different characteristics of the product items. 
In order to exploit the persona information, we propose a framework based on adversarial variational auto-encoders (aVAE) for persona modeling from the historical tips and reviews of users and items. The latent variables from aVAE are regarded as persona embeddings. Besides representing persona using the latent embeddings, we design a persona memory for storing the persona related words for users and items. Pointer Network is used to retrieve persona wordings from the memory when generating tips. Moreover, the persona embeddings are used as latent factors by a rating prediction component to predict the sentiment of a user over an item.
Finally, the persona embeddings and the sentiment information are incorporated into a recurrent neural networks based tips generation component. Extensive experimental results are reported and discussed to elaborate the peculiarities of our framework.
\end{abstract}

%
%



\maketitle


\section{Introduction}
\label{section1}

Tips, specifically defined by Yelp\footnote{https://www.yelp-support.com/article/What-are-tips}, as a compacted and concise form of reviews, have unique advantages for helping users get a quick insight over an item. 
Conventional reviews are extensively studied for rating prediction \cite{mcauley2013hidden,wang2011collaborative} and review generation \cite{tang2016context,dong2017learning,ni2017estimating,yao2017automated}, while tips are paid relatively less attention. In \cite{li2017neural}, the tip information was explored for tips generation and rating prediction for the first time. The rationality for performing the joint task can be attributed to ``writing some tips and giving a numerical rating are two facets of a user's product assessment action, expressing the user experience and feelings''. Moreover, compared with reviews, tips are likely more consistent with the rating score with respect to sentiment tendency, because of its intrinsic form, i.e. compacted and concise.

\begin{figure}[!t]
	\centering
	\begin{subtable}{.99\linewidth}
		\centering
		\small
		\begin{tabular}{@{~}p{0.82\columnwidth} | @{~}c@{~}}
			\hline
			\textbf{\hspace{2.8cm}  Tips} & \textbf{Rating}\\
			\hline
			(1)\hspace{1mm} Great fit and finish for shower. & 5\\
			(2)\hspace{1mm} I selected this radio for myself several years ago and i have found that all claims for it are true.  & 5\\
			(3)\hspace{1mm} If your looking for a radio for your shower then look no further.  & 5\\
			(4)\hspace{1mm} Easy to set up stations.  & 5\\
			(5)\hspace{1mm} Excellent design and quality construction.   & 5\\
			(6)\hspace{1mm} First one lasted years just bought another one.  & 5\\
			\hline
		\end{tabular}
		\caption{\label{fig:front-a} Tips for the item ``Sony Weather Band Shower Radio''.}
	\end{subtable}
	\\
	\vspace{2mm}
	\begin{subtable}{.99\linewidth}
		\centering
		\small
		\begin{tabular}{@{~}p{0.82\columnwidth} |@{~} c@{~}}
			\hline
			\textbf{\hspace{2.8cm}  Tips} & \textbf{Rating}\\
			\hline
			(1)\hspace{1mm} Works perfectly in my msi wind. & 5\\
			(2)\hspace{1mm} Perfect size for a home office. & 5\\
			(3)\hspace{1mm} Excellent player for price. & 5\\
			(4)\hspace{1mm} Wonderful docking speaker with full sound. & 4\\
			(5)\hspace{1mm} I like it when it not dropping the signal. & 4 \\
			(6)\hspace{1mm} Works fine in a pinch. & 3\\
			(7)\hspace{1mm} Piece of crap do bother. & 1 \\
			(8)\hspace{1mm} Revised star piece of crap. & 1 \\
			\hline
		\end{tabular}
		\caption{\label{fig:front-b}Tips for different items written by a particular user.}
	\end{subtable} 
	
	\caption{\label{fig:front}
		Example of tips. 
	}
\end{figure}

In this paper, we investigate another dimension, namely user persona, which is plausibly helpful for the task of tips generation and has not be considered in the previous work \cite{li2017neural}. Here the term ``\textbf{persona}'' denotes the characteristics of the written text by users such as wording and style.
Figure~\ref{fig:front-a} shows some tips for a shower radio from different users. \footnote{https://www.amazon.com/sony-icf-s79v-weather-shower-radio/dp/b00000dm9w} These tips clearly show different styles, although all of them have the same ratings. Some users (e.g. 1, 4 and 5) prefer short sentences and direct wordings such as ``great'', ``easy'', and ``excellent'' to describe the product quality and their experience directly. On the other hand, some users (e.g. 2, 3, and 6) share their experience indirectly by talking about some facts with longer sentences. Therefore, different users indeed have different ``persona'' style when writing tips.
Figure~\ref{fig:front-b} shows a few tips with different ratings from the same user for different items, we can observe that the user prefers short sentences, and moreover he has his own style (i.e. preferred vocabulary) for writing tips of different sentiments/ratings (e.g.  ``perfect'' and ``excellent'' for high rating tips, and ``piece of crap'' for low rating tips).

Intuitively, the quality of abstractive tips generation can be improved if the model considers the user persona information when conducting the text generation. To do so, in this paper, we investigate an approach called \textbf{P}ersona-\textbf{A}ware \textbf{T}ips \textbf{G}eneration (\textbf{PATG}). There are two main challenges for the design of PATG: (1) How to capture and represent the persona information; (2) How to integrate the sentiment signal with the persona information to jointly control the style and the sentiment of the generated tips.

We distill persona information from all the historical tips and reviews of a user into the form of \textbf{Persona Embeddings}. Then the persona embeddings can be directly incorporated into the tips generation component as context information. Specifically, we employ variational auto-encoders (VAEs) \cite{kingma2013auto} (which show strong capability in modeling latent random variables \cite{li2017collaborative,li2017salience}) to conduct persona modeling, and we regard the latent variables of VAEs as the persona embeddings. In the context of user behaviour analysis of online retailing, another indispensable party is the product item. Items also have their own intrinsic characteristics, such as product category in a coarser granularity, or specific features in a finer granularity. In this work, we personify the items and enable modeling them with ``item persona'', similar to modeling users. Besides distilling persona information using VAEs, we also design an external \textbf{Persona Memory} for the framework to store the persona related words for the current user and item. Pointer Networks \cite{vinyals2015pointer} is used to retrieve appropriate words from the persona memory for generating tips.

Another obvious signal from the examples in Figure \ref{fig:front} is that the sentiment related wording is also bound to the sentiment ratings. For example, positive expressions are only used in tips with high ratings, no matter what persona the users have. To explore this signal for generating more accurate tips, our framework includes an auxiliary component: rating prediction with the information of users and items used for tips generation. The intuition is that if we can predict the rating of a user on an item accurately, the same input information should provide rich, if not complete, information for generating a tip satisfying that rating. Thus, in order to control the sentiment of the generated tips, we design a rating prediction component. The distilled persona embeddings are regarded as latent factors for users/items, and fed into the rating prediction component for detecting sentiment. A vectorization process is conducted on the predicted rating values and then the rating vectors are incorporated into the tips generation component as context information to control the sentiment of the generated tips. 

The main contributions of our framework are summarized below:
\begin{itemize}
	\item We develop a framework that tackles the task of persona-aware tips generation, where persona information, such as writing style and vocabulary preference, is considered for the first time to conduct the tips generation.  
	
	\item In order to exploit the persona information, we design an adversarial variational auto-encoders (aVAE) based approach for persona modeling for users and items, i.e. generating persona embeddings. We employ an external memory based Pointer Networks to conduct the memory reading to retrieve more accurate persona information.
	
	\item In order to control the sentiment of the generated tips, we tightly couple an auxiliary component of rating prediction with the tips generation component. The distilled persona embeddings are used as latent factors of users and items for the sentiment rating prediction. 
	
	\item Experimental results show that our framework achieves better performance than the state-of-the-art models on tips generation. Moreover, an additional observation is that the persona information can improve the performance of the auxiliary task, i.e. rating prediction.
\end{itemize}

\section{Framework Description}
\label{section3}

\begin{figure*}[!t]
	\centering
	\includegraphics[width=1.99\columnwidth]{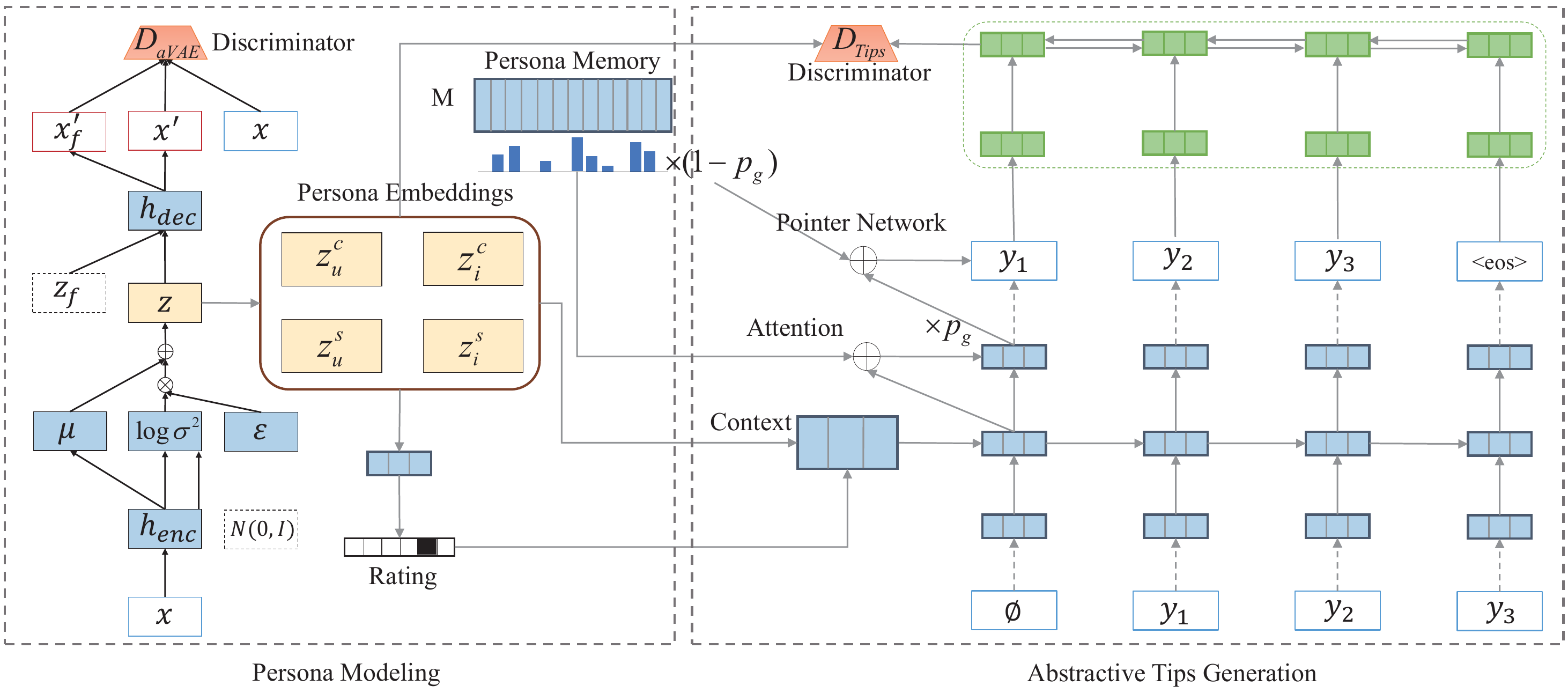}
	\caption{\label{fig:framework}
		Our proposed framework for persona-aware abstractive tips generation.
	}
\end{figure*}

\subsection{Overview}
The data consists of users, items, ratings, review content, and tips.
We denote the whole training corpus by  $\mathcal{X} = \{\mathcal{U}, \mathcal{I}, \mathcal{R}, \mathcal{C}, \mathcal{S} \}$, where $\mathcal{U}$ and $\mathcal{I}$ are the sets of users and items respectively, $\mathcal{R}$ is the set of ratings, $\mathcal{C}$ is the set of review documents, and $\mathcal{S}$ is the set of tips texts. We use  $\mathcal{C}_u$ and $\mathcal{S}_u$ to denote all the historical reviews and tips respectively of the user $u$. For a quick reference, Table~\ref{tbl:notations} lists all notations used in our paper.

\begin{table}[!t]
	\caption{Glossary.}
	\label{tbl:notations}
	\centering
	\begin{tabular}{p{1.3cm} p{4cm}}
		\hline
		Symbol & Description \\
		\hline
		$\mathcal{X}$ & training set\\
		$\mathcal{V}$ & vocabulary\\
		$\mathcal{U}$  & set of users \\
		$\mathcal{I}$  & set of items \\
		$\mathcal{R}$  & set of ratings \\
		$\mathcal{C}$  & set of reviews \\
		$\mathcal{S}$  & set of tips \\
		$\mathcal{C}_u$  & historical reviews for user $u$ \\
		$\mathcal{S}_u$  & historical tips for user $u$ \\
		$\mathbf{M}$  & external memory \\
		$\mathbf{Z}$  & persona embeddings \\
		$\mathbf{E}$  & word embeddings \\
		$\mathbf{H}$  & neural hidden states \\
		$\mathbf{W}$  & mapping matrix\\
		$\mathbf{b}$  & bias item \\
		$\mathbf{\Theta}$  & set of neural parameters\\
		$r_{u,i}$  & rating of user $u$ to item $j$ \\
		$\sigma$ & sigmoid function \\
		$\varsigma $ & softmax function \\
		$relu$ & rectified linear unit \\
		$tanh$ &  hyperbolic tangent function \\
		\hline
	\end{tabular}
\end{table}

As shown in Figure~\ref{fig:framework}, our framework contains two major modules: persona modeling on the left and abstractive tips generation on the right.
For modeling persona, our framework leverages the tips and reviews from each individual user or written by multiple users for the same item. Take the historical tips $\mathcal{S}_u$ of the user $u$ as an example, we represent them using bag-of-words (BoWs) vectors $\mathbf{x}_u^s$. Then we feed $\mathbf{x}_u^s$ into the adversarial variational auto-encoders (aVAE$^s$) and obtain the persona embedding $\mathbf{z}_u^s$ for the user $u$. For the item $i$, we can also conduct similar persona modeling based on the historical tips $\mathcal{S}_i$ written by different users, and the obtained persona embedding is denoted as $\mathbf{z}_i^s$. The purpose of persona modeling for the item $i$ is that when conducting tips generation for the user $u$, the model will also consider the tips from other users having similar interests with $u$, since they will disclose more characteristics of the item. We call this phenomenon \textit{personalized collaborative influence}. We also distill persona information from reviews with another aVAE model (aVAE$^c$) to map the historical reviews $\mathcal{C}_u$ and $\mathcal{C}_i$ to persona embeddings $\mathbf{z}_u^c$ and $\mathbf{z}_i^c$ for the user $u$ and the item $i$ respectively.

We design an external persona memory $\mathbf{M}$ for storing the persona related words for the current user and item which will be utilized in abstractive tips generation. In order to control the sentiment of the generated tips, the distilled persona embeddings are used as latent factors for users and items and are fed into a multilayer perceptron (MLP) based neural network component to predict the rating $r$. Then we transform $r$ to a one-hot vector $\mathbf{r}$ which will be used as the sentiment controller when conducting the tips generation. For the step of tips generation, we design a sequence decoding model based on a neural network of Gated Recurrent Units (GRUs) \cite{cho2014learning}. Importantly, the persona embeddings and the rating vector are combined to construct a context vector which plays a significant role in the abstractive tips generation. In addition, Pointer Networks is used to retrieve relevant words from the persona memory $\mathbf{M}$, with a gate $p_g$ to control the source of the next output word.

\subsection{Persona Modeling}

\subsubsection{\textbf{Persona Embedding Learning}}

The target of persona modeling is to distill the persona information from the users' historical tips and reviews. Some previous works in recommendation systems \cite{wang2011collaborative,mcauley2013hidden,rensocial2017} employ topic modeling methods such as Latent Dirichlet Allocation (LDA) \cite{blei2003latent}  or its variants to analyze the text corpus and use the latent topic distribution to represent each document. 
Considering the fact that our tips generation component is based on neural networks, existing topic modeling paradigms cannot be incorporated into our framework in an elegant manner. Instead, we employ the variational auto-encoders (VAEs) \cite{kingma2013auto} for detecting the latent topics with neural modeling paradigm \cite{card2017neural}. VAEs consists of two parts: inference (variational-encoder) and generation (variational-decoder). Recall that the dictionary is $V$.
For historical tips based persona modeling, the input are the BoWs vectors $\mathbf{x}_u^s \in \mathbb{R}^{|V|}$ and $\mathbf{x}_i^s \in \mathbb{R}^{|V|}$ for the user $u$ and the item $i$ respectively. For convenience, we will use $\mathbf{x}$ to represent them in this section.
As shown in the left part of Figure~\ref{fig:framework}, for each input BoWs vector  $\mathbf{x}$, the variational-encoder can map it to a latent variable $\mathbf{z} \in {\mathbb{R}^K}$, which can be used to generate a new variable $\mathbf{x}^\prime$  via the variational-decoder component to reconstruct the original term vector. The target is to maximize the probability of each $\mathbf{x}$ in the dataset based on the generation process according to:
\begin{equation}
	{p}(\mathbf{x}) = \int {{p}(\mathbf{x}|\mathbf{z}){p }(\mathbf{z})d\mathbf{z}}
\end{equation}
For the purpose of solving the intractable integral of the marginal likelihood, a model $q(\mathbf{z}|\mathbf{x})$ is introduced as the approximation to the intractable of the true posterior $p(\mathbf{z}|\mathbf{x})$. The aim of optimization is to reduce the Kulllback-Leibler divergence (KL) between $q(\mathbf{z}|\mathbf{x})$ and $p(\mathbf{z}|\mathbf{x})$ by maximizing the variational lower bound $\mathcal{L}_{VAE}$:
\begin{equation}
	\mathcal{L}_{VAE} = {\mathbb{E}_{{q}(\mathbf{z}|\mathbf{x})}}[\log {p}(\mathbf{x}|\mathbf{z})] - {D_{KL}}[{q}(\mathbf{z}|\mathbf{x})\|{p}(\mathbf{z})]
\end{equation}
In order to differentiate and optimize the lower bound $\mathcal{L}_{VAE}$, following the core idea of VAEs, we use a neural network framework for the encoder $q(\mathbf{z}|\mathbf{x})$ for better approximation.
Similar to previous works \cite{kingma2013auto}, we assume that both the prior and posterior of the latent variables are Gaussian, i.e.,  $p(\mathbf{z}) = \mathcal{N}(0, \mathbf{I})$ and $q(\mathbf{z}|\mathbf{x}) = \mathcal{N}(\mathbf{z}; \boldsymbol{\mu}, \boldsymbol{\sigma}^2\mathbf{I})$, where  $\boldsymbol{\mu}$ and $\boldsymbol{\sigma}$ denote the variational mean and standard deviation respectively, which can be calculated with a multilayer perceptron (MLP). 
Precisely, given the BoWs vector $\mathbf{x}$ of the historical tips, we first project it to a hidden space:
\begin{equation}
	{\mathbf{h}_{enc}} = relu({\mathbf{W}_{xh}}\mathbf{x} + {\mathbf{b}_{xh}})
\end{equation}
where $\mathbf{h}_{enc} \in \mathbb{R}^{d_h}$, $\mathbf{W}_{xh}$ and $\mathbf{b}_{xh}$ are the neural parameters. $relu(\mathbf{x}) = max(0, \mathbf{x})$ is the activation function.
Then the Gaussian parameters $\boldsymbol{\mu} \in \mathbb{R}^K$ and $\boldsymbol{\sigma} \in \mathbb{R}^K$ can be obtained via a linear transformation based on  $\mathbf{h}_{enc}$:
\begin{equation}
	\begin{array}{l}
		\boldsymbol{\mu } = {\mathbf{W}_{h\mu }}{\mathbf{h}_{enc}} + {\mathbf{b}_{h\mu }}\\
		\log ({\boldsymbol{\sigma} ^2}) = {\mathbf{W}_{h\sigma }}{\mathbf{h}_{enc}} + {\mathbf{b}_{h\sigma }}\\
	\end{array}
\end{equation}
In order to make the sampling operation differentiable, the latent variable $\mathbf{z} \in \mathbb{R}^K$ can be calculated using the reparameterization trick:
\begin{equation}
	\label{e:ptrick}
	\begin{array}{l}
		\boldsymbol{\varepsilon}  \sim \mathcal{N}(0, \mathbf{I}), \ \
		\mathbf{z} = \boldsymbol{\mu }  + \boldsymbol{\sigma}  \otimes \boldsymbol{\varepsilon} \\
	\end{array}
\end{equation}
where $\boldsymbol{\varepsilon} \in \mathbb{R}^K$ is an auxiliary noise variable. This is the encoding process, and we denote all the parameters of this state as $\Theta_{Enc}$.

Given the latent variable $\mathbf{z}$, a new vector $\mathbf{x'}$ is generated via the conditional distribution ${p}(\mathbf{x}|\mathbf{z})$ according to the variational-decoder: 
\begin{equation}
	\begin{array}{l}
		{\mathbf{h}_{dec}} = relu({\mathbf{W}_{zh}}\mathbf{z} + {\mathbf{b}_{zh}})
	\end{array}
\end{equation}
\begin{equation}
	\begin{array}{l}
		{\mathbf{x}^\prime} = \sigma({\mathbf{W}_{hx}}{\mathbf{h}_{dec}} + {\mathbf{b}_{hx}})
	\end{array}
\end{equation}
We denote all the parameters in the decoding stage using $\Theta_{Dec}$. Finally, based on the reparameterization trick in Equation~\ref{e:ptrick},  we can get the analytical representation of $\mathcal{L}_{VAE}$:
\begin{equation}
	\begin{array}{l}
		\log p(\mathbf{x}|\mathbf{z}) = \sum\limits_{i = 1}^{|V|} {{x_i}\log x_i^\prime + (1 - {x_i}) \cdot \log (1 - x_i^\prime)} \\
		- {D_{KL}}[{q }(\mathbf{z}|\mathbf{x})\|{p }(\mathbf{z})]{\rm{ = }}\frac{1}{2}\sum\limits_{i = 1}^K {(1 + \log (\sigma _i^2) - \mu _i^2 - \sigma _i^2)}
	\end{array}
	\label{eq:vae_loss}
\end{equation}
For presentation clarity, we let:
\begin{equation}
	\begin{array}{l}
		\mathcal{L}_{Rec} = -\log p(\mathbf{x}|\mathbf{z}) \\
	\mathcal{L}_{KL} = {D_{KL}}[{q }(\mathbf{z}|\mathbf{x})\|{p }(\mathbf{z})]
	\end{array}
\end{equation}
and both of them need to be minimized.

We wish to employ the latent variable $\mathbf{z}$ as the distilled persona embeddings. So the quality of $\mathbf{z}$ will affect the performance of tips generation. Some previous works \cite{goyal2017z,zhao2017learning,mescheder2017adversarial} have also shown that the performance of $\mathbf{z}$ is likely to be disturbed during the training procedure, especially when combining VAEs with the RNN based text generation framework. In order to enhance the performance of the typical VAEs, inspired by the ideas in \cite{goodfellow2014generative} and \cite{larsen2016autoencoding}, we employ the adversarial strategy for the training of VAEs. Generally, we design a discriminator network $D_{aVAE}$ with a vector $\mathbf{\tilde x}$ as input, and the target is to recognize if $\mathbf{\tilde x}$ is from the true data $\mathbf{X}$ or from the generated samples $\mathbf{X^\prime}$ by VAEs.  VAEs will ``fool'' the discriminator $D_{aVAE}$ by trying the best to produce high quality latent variables $\mathbf{z}$ as well as the generated sample $\mathbf{x^\prime}$. Then the minimax game between the VAEs and the discriminator can be formulated as follows: 
\begin{equation}
	\begin{split}
		\mathop {\min }\limits_{VAEs} \mathop {\max }\limits_{{D_{aVAE}}} &{\mathbb{E}_{\mathbf{x} \sim {p_{data}}(\mathbf{x})}}[\log {D_{aVAE}}(\mathbf{x})]\\
		&+ {\mathbb{E}_{\mathbf{z} \sim p(\mathbf{z}|\mathbf{x})}}[\log (1 - {D_{aVAE}}(VA{E_{Dec}}(\mathbf{z})))]\\
		&+ {\mathbb{E}_{{\mathbf{z}_f} \sim p(\mathbf{z})}}[\log (1 - {D_{aVAE}}(VA{E_{Dec}}({\mathbf{z}_f})))]
	\end{split}
	\label{eq:loss_mima}
\end{equation}
where $VAE_{Dec}$ is the decoder component of the VAEs model. $\mathbf{z}$ is the latent variable from VAEs, and $\mathbf{z}_f$ is sampled from the prior distribution of $\mathbf{z}$.

For the design of the discriminator $D_{aVAE}$, we simply use a multilayer perceptron to process the data.
\begin{equation}
	\begin{split}
		{\mathbf{h}^{D_v}} = tanh({\mathbf{W}^{D_{v}}_{xh}}\mathbf{\tilde{x}} + {\mathbf{b}^{D_{v}}_{xh}})
		\\
		{y^{D_v}} = \sigma({\mathbf{W}^{D_{v}}_{hy}}{\mathbf{h}^{D_v}} + {b^{D_{v}}_{hy}})
	\end{split}
\end{equation}
where $\mathbf{W}^{D_v}_{xh} \in \mathbb{R}^{d_h \times |V|} $, $\mathbf{W}^{D_v}_{hy} \in \mathbb{R}^{1 \times d_h} $, $\mathbf{b}^{D_v}_{xh} \in \mathbb{R}^{d_h}$, and $b^{D_v}_{hy} \in \mathbb{R}$.
The output $y^{D_v}$ is a real value in the range of $[0,1]$ and the value 1 means that the sample $\tilde{\mathbf{x}}$ is from the true data. We denote all the parameters in $D_{aVAE}$ using $\Theta_{D_v}$. The optimization objective to be maximized for $D_{aVAE}$ is formulated as:
\begin{equation}
	\begin{split}
		{\mathcal{L}_{{D_{aVAE}}}} = &\log ({D_{aVAE}}(\mathbf{x})) \\
		&+ \log (1 - {D_{aVAE}}(VA{E_{Dec}}(VA{E_{Enc}}(\mathbf{x})))) \\
		&+ \log (1 - {D_{aVAE}}(VA{E_{Dec}}({\mathbf{z}_f})))
	\end{split}
	\label{eq:loss_d}
\end{equation}
Then the parameters $\Theta_{D_v}$ are updated using gradient methods:
\begin{equation}
	{\Theta _{{D_v}}} \leftarrow {\Theta _{{D_v}}} - {\nabla _{{\Theta _{{D_v}}}}}( - {\mathcal{L}_{{D_{aVAE}}}})
\end{equation}

Conditioned on the aVAE framework, we will conduct the parameter learning for VAEs Encoder, VAEs Decoder, and discriminator $D_{aVAE}$ using different loss functions respectively. Encoder transforms the input $\mathbf{X}$ to the persona embeddings $\mathbf{Z}$. On one side,  $\mathbf{Z}$ are used to reconstruct the original input. On the other side,  $\mathbf{Z}$ are used to conduct the persona-aware tips generation. So the loss signals from both the aVAE and the tips generation framework are used to conduct the optimization for $\Theta_{Enc}$:
\begin{equation}
	{\Theta _{Enc}} \leftarrow {\Theta _{Enc}} - {\nabla _{{\Theta _{Enc}}}}({\mathcal{L}_{KL}} + {\mathcal{L}_{Rec}} + \mathcal{L}_{{D_{aVAE}}}^{z} + \mathcal{L}_{Tips} )
\end{equation}
where $\mathcal{L}_{KL}$ and $\mathcal{L}_{Rec}$ are the KL divergence and reconstruction loss from Equation~\ref{eq:vae_loss}. $\mathcal{L}_{Tips}$ is the loss signal from the tips generation component. $\mathcal{L}_{{D_{aVAE}}}^{z}$ is the output of $D_{aVAE}$:
\begin{equation}
	\mathcal{L}_{{D_{aVAE}}}^{z} = -\log ({D_{aVAE}}(VA{E_{Dec}}(VA{E_{Enc}}(\mathbf{x}))))
\end{equation}

For the parameter optimization of VAEs Decoder, we use $\mathcal{L}_{Rec}$, $\mathcal{L}_{{D_{aVAE}}}$, $\mathcal{L}_{Tips}$ as the loss signals:
\begin{equation}
	{\Theta _{Dec}} \leftarrow {\Theta _{Dec}} - {\nabla _{{\Theta _{Dec}}}}({\mathcal{L}_{{\mathop{\rm Re}\nolimits} c}} + {\mathcal{L}_{{D_{aVAE}}}} + {\mathcal{L}_{Tips}})
\end{equation}
Finally, the training procedure of aVAE model is shown in Algorithm~\ref{alg:avae}. 

\begin{algorithm}[!t]
	\caption{Persona embedding learning.}
	\label{alg:avae}
	\begin{algorithmic}[1]
		\REQUIRE BoWs vectors of historical tips and reviews $\mathbf{X}$.
		\ENSURE The persona embeddings $\mathbf{Z}$.
		\STATE Initialize  $\Theta_{Enc}, \Theta_{Dec}, \Theta_{D_v}$;
		\WHILE{not converged}
		\STATE Draw $\mathbf{x}$ from $p_{data}$.
		\STATE Draw $\mathbf{z}_f$ from prior $p(\mathbf{z})$.
		\STATE $\mathbf{z} = VAE_{Enc}(\mathbf{x})$
		\STATE $\mathbf{x}^\prime = VAE_{Dec}(\mathbf{z})$
		\STATE $\mathbf{x}^\prime_f = VAE_{Dec}(\mathbf{z}_f)$
		\STATE Get $\mathcal{L}_{Rec}$,  $\mathcal{L}_{KL}$, $\mathcal{L}_{{D_{aVAE}}}$ according to Equation~\ref{eq:vae_loss} and \ref{eq:loss_d}.
		\STATE Get $\mathcal{L}_{Tips}$ from tips generation.
		\STATE Update parameters using gradient methods:\\
		${\Theta _{Enc}} \leftarrow {\Theta _{Enc}} - {\nabla _{{\Theta _{Enc}}}}({\mathcal{L}_{KL}} + {\mathcal{L}_{Rec}} + \mathcal{L}_{{D_{aVAE}}}^{z} + \mathcal{L}_{Tips} )$\\
		${\Theta _{Dec}} \leftarrow {\Theta _{Dec}} - {\nabla _{{\Theta _{Dec}}}}({\mathcal{L}_{{\mathop{\rm Re}\nolimits} c}} + {\mathcal{L}_{{D_{aVAE}}}} + {\mathcal{L}_{Tips}})$\\
		${\Theta _{{D_v}}} \leftarrow {\Theta _{{D_v}}} - {\nabla _{{\Theta _{{D_v}}}}}( - {L_{{D_{aVAE}}}})$
		\ENDWHILE
		\RETURN $\mathbf{z}$.
	\end{algorithmic}
\end{algorithm}

Feeding the historical reviews and tips representations ($\mathbf{x}_u^c$, $\mathbf{x}_i^c$, $\mathbf{x}_u^s$, and $\mathbf{x}_i^s$) into $aVAE^c$ (for reviews) and $aVAE^s$ (for tips) respectively, we can obtain four persona embeddings $\mathbf{z}_u^c$, $\mathbf{z}_i^c$, $\mathbf{z}_u^s$, and $\mathbf{z}_i^s$. These persona embeddings will be integrated into the rating prediction component and the tips generation component later.

\subsubsection{\textbf{Sentiment and Rating Modeling}}
We regard the persona embeddings as the latent factors of users and items, and feed them into a multilayer perceptron to conduct the rating prediction. The predicted ratings  will be used to control the sentiment of the generated tips.

Specifically, we first map the persona embeddings to a hidden space:
\begin{equation}
	{\mathbf{h}^r} = tanh (\mathbf{W}_{uch}^r\mathbf{z}_u^c + \mathbf{W}_{ich}^r\mathbf{z}_i^c +  \mathbf{W}_{ush}^r\mathbf{z}_u^s + \mathbf{W}_{ish}^r\mathbf{z}_i^s +  \mathbf{b}_h^r)
	\label{eq:hr}
\end{equation}
where $\{\mathbf{W}_{uch}^r,  \mathbf{W}_{ich}^r , \mathbf{W}_{ush}^r, \mathbf{W}_{ish}^r\} \in \mathbb{R}^{d_h \times k}$ are the mapping matrices. $\mathbf{b}_h^r \in \mathbb{R}^{d_h}$ is the bias term. $tanh$ is the  hyperbolic tangent activation function.
The superscript $r$ refers to variables related to the rating prediction component.
For better performance, we can add more layers of non-linear transformations into our model:
\begin{equation}
	{\mathbf{h}^r_l} = \sigma (\mathbf{W}_{h{h_l}}^r\mathbf{h}^r_{l-1} + \mathbf{b}_{{h_l}}^r)
\end{equation}
where $\mathbf{W}_{h{h_l}}^r \in \mathbb{R}^{d_h \times d_h}$ is the mapping matrix for the variables in the hidden layers. $l$ is the index of a hidden layer.
Assume that $\mathbf{h}^r_L$ is the output of the last hidden layer.
The output layer transforms $\mathbf{h}^r_L$ into a real-valued rating $\hat r$:
\begin{equation}
	{\hat r} = \mathbf{W}_{hr}^r\mathbf{h}^r_{L} + b^r
	\label{eq:pred_r}
\end{equation}
where $\mathbf{W}_{hr}^r \in \mathbb{R}^{1 \times d_h}$ and $b^r \in \mathbb{R}$.
We formulate the optimization of the parameters $\Theta_{r}$ as a regression problem and  the loss function is formulated as:
\begin{equation}
	{\mathcal{L}^r} = \frac{1}{{2\left| \mathcal{X} \right|}}{\sum\limits_{u\in \mathcal{U},i \in \mathcal{I}} {({{\hat r}_{u,i}} - {r_{u,i}})} ^2}
	\label{eq:lossr}
\end{equation}
where $\mathcal{X}$ represents the training set. $r_{u,i}$ is the ground truth rating assigned by the user $u$ to the item $i$.

The predicted rating is a real value, not a vector, for example, $\hat r_{u,i} = 4.321$. In order to incorporate the rating information into the tips generation component, we cast it into an integer $4$, and add a vectorization process to obtain the vector representation of rating $\hat r_{u,i}$. If the rating range is $[0, 5]$, we will get the rating vector $\mathbf{\hat r}_{u,i} = (0, 0, 0, 0, 1, 0)^T$.

\subsubsection{\textbf{External Persona Memory}}
\label{sec:pm}
In addition to represent persona information using the latent embeddings from aVAE, we design an external persona memory for directly storing the persona related words for both the current user $u$ and the current item $i$. 
To build the memory, we first collect all the words for the current user $u$ and the current item $t$ from their historical tips. We add a filtering process to remove the stop-words and the low-frequency words.
Then we get a local vocabulary storing the indices of the persona words. Recall that we have a global word embedding $\mathbf{E}$. Then we can get a sub-matrix from $\mathbf{E}$ according to the word indices. We regard this sub-matrix as persona memory.
We employ Pointer Networks to retrieve persona information from the memory when generating tips. The details are described in Section~\ref{sec:pn}.

\subsection{Abstractive Tips Generation}

\subsubsection{\textbf{Overview of Tips Generation}}
The right part of Figure~\ref{fig:framework} depicts our tips generation model. The basic element is a RNN based sequence modeling component. Pointer Networks (attention modeling and copy mechanism) is introduced to conduct the memory reading.
Context information plays an important role in the task of text generation. We combine the persona embeddings and the sentiment information as the context information and construct the context vector which can control the tips text generation.
At the training state, we also design a discriminator $D_{Tips}$ to assess the quality of the generated tips. The assess value will be propagated to the RNN models to assist the parameter learning.
At the operational or testing stage, we use a beam search algorithm \cite{koehn2004pharaoh} for decoding and generating the best tips given a trained model.

\subsubsection{\textbf{Sequence Modeling}}
Assume that $\mathbf{h}_t^s$ is the sequence hidden state at the time $t$. It depends on the input at the time $t$ and the previous hidden state  $\mathbf{h}_{t-1}^s$:
\begin{equation}
	\mathbf{h}_{t}^s =  f(\mathbf{h}_{t-1}^s, s_{t})
	\label{eq:rnn-h}
\end{equation}
$f(\cdot)$ can be the vanilla RNN,  Long Short-Term Memory (LSTM) \cite{hochreiter1997long}, or  Gated Recurrent Unit (GRU) \cite{cho2014learning}. Considering that GRU has comparable performance but with less parameters and more efficient computation, we employ GRU as the basic model in our sequence modeling framework.
In the case of GRU, the state updates are processed according to the following operations:
\begin{equation}
	\begin{array}{l}
		\mathbf{r}_t^s = \sigma (\mathbf{W}_{sr}^s\mathbf{s}_t + \mathbf{W}_{hr}^s\mathbf{h}_{t - 1}^s + \mathbf{b}_r^s)\\
		\mathbf{z}_t^s = \sigma (\mathbf{W}_{sz}^s\mathbf{s}_t + \mathbf{W}_{hz}^s\mathbf{h}_{t - 1}^s + \mathbf{b}_z^s)\\
		\mathbf{g}_t^s = \tanh (\mathbf{W}_{sh}^s\mathbf{s}_t + \mathbf{W}_{hh}^s(\mathbf{r}_t^s \odot \mathbf{h}_{t - 1}^s) + \mathbf{b}_h^s)\\
		\mathbf{h}_t^s = \mathbf{z}_t^s \odot \mathbf{h}_{t - 1}^s + (1 - \mathbf{z}_t^s) \odot \mathbf{g}_t^s
	\end{array}
	\label{eq:gru}
\end{equation}
where $\mathbf{s}_t \in \mathbf{E}$ is the embedding vector for the word $s_t$ of the tips and the vector is also learnt from our framework.
$\mathbf{r}_t^s$ is the reset gate, $\mathbf{z}_t^s$ is the update gate.
$\odot$ denotes element-wise multiplication.

In order to conduct the persona-aware tips generation, we combine all the persona embeddings and the sentiment information as the context information and construct the context vector. Specifically, we initialize the hidden state $\mathbf{h}_0$ using the persona embeddings and the sentiment information:
\begin{equation}
	\mathbf{h}_{0}^s = tanh(\mathbf{W}_{uch}^s\mathbf{z_u^c} + \mathbf{W}_{ich}^s\mathbf{z_i^c} + \mathbf{W}_{ush}^s\mathbf{z_u^s} + \mathbf{W}_{ish}^s\mathbf{z_i^s} + \mathbf{W}_{rh}^s\mathbf{\hat r} +  \mathbf{b}_h^s)
\end{equation}
where $\{\mathbf{z_*^*}\}$ are the persona embeddings. $\mathbf{\hat r}$ is the vectorization for the predicted rating $\hat r$. $\mathbf{W}$ and $\mathbf{b}$ are the neural parameters.

After getting all the sequence hidden states based on GRU, we feed them to the final output layer to predict the word sequence in tips.
\begin{equation}
	\mathbf{\hat s}_{t+1} = \varsigma (\mathbf{W}_{hs}^s\mathbf{h}_t^s  + \mathbf{b}^s)
	\label{eq:rnn-o}
\end{equation}
where $\mathbf{W}_{hs}^s \in \mathbb{R}^{d \times |\mathcal{V}|}$ and $\mathbf{b}^s \in \mathbb{R}^{|\mathcal{V}|}$. $\varsigma(\cdot)$ is the softmax function. Then the word with the largest probability is the decoding result for the step $t+1$:
\begin{equation}
	{w_{t + 1}^*} = \mathop {\arg\max }\limits_{{w_i} \in \mathcal{V}} \mathbf{\hat s}_{t + 1}^{ (w_i) }
\end{equation}

At the training stage, we use negative log-likelihood (NLL) as the loss function, where $I_w$ is the vocabulary index of the  word $w$:
\begin{equation}
	{\mathcal{L}_{Tips}} =  - \sum\limits_{w \in Tips} {\log {\mathbf{\hat s}^{(I_w)}}}
	\label{eq:losss}
\end{equation}
Note that $\mathcal{L}_{Tips}$ is also used in the persona modeling component to train the aVAE models.

At the testing stage, given a trained model, we employ the beam search algorithm \cite{koehn2004pharaoh} to find the best sequence $S^*$ having the maximum log-likelihood.
\begin{equation}
	{S^*} = \mathop {\arg \max }\limits_{S \in \mathcal{S}} \sum\limits_{w \in S} {\log {\mathbf{\hat s}^{(I_w)}}}
\end{equation}

\subsubsection{\textbf{Exploiting Persona Memory}}
\label{sec:pn}
Recall that in Section~\ref{sec:pm}, we build a local personal vocabulary $V_{ui}$ for the user $u$ and the item $i$. The persona memory $\mathbf{M}_{ui}$ is extracted from the word embedding $\mathbf{E}$ using the word indices in $V_{ui}$. 
Inspired by \cite{bahdanau2014neural}, we exploit the idea of attention modeling to conduct the addressing and reading operations on the memory $\mathbf{M}_{ui}$. We can obtain the GRU hidden state $\mathbf{h}_t^s$ according to Equation~(\ref{eq:rnn-h}).
Then the attention weights at the time step $t$ are calculated based on the relationship between $\mathbf{h}_t^s$ with all the word embeddings in $\mathbf{M}_{ui}$. Let $a_{i,j}$ be the attention weight between $\mathbf{h}_i^{s}$ and $\mathbf{m}_j$, which can be calculated using:
\begin{equation}
	\begin{split}
		{a_{i,j}} &= \frac{{\exp ({e_{i,j}})}}{{\sum\nolimits_{j' = 1}^{{|V_{ui}|}} {\exp ({e_{i,j'}})} }}\\
		{e_{i,j}} &= {{\mathbf{v}_a}^T}\tanh (\mathbf{W}_{hh}^s\mathbf{h}_i^{{s}} + \mathbf{W}_{hh}^m\mathbf{m}_j + {\mathbf{b}_a})
	\end{split}
\end{equation}
where $\mathbf{W}_{hh}^s \in \mathbb{R}^{d_h \times d_h}$, $\mathbf{W}_{hh}^m \in \mathbb{R}^{d_w \times d_h}$, $\mathbf{b}_a \in \mathbb{R}^{d_h}$, and $\mathbf{v}_a \in \mathbb{R}^{d_h}$.
The attention context is obtained by the weighted linear combination of all the word embeddings in $\mathbf{M}_{ui}$:
\begin{equation}
	{\mathbf{c}_t} = \sum\nolimits_{j' = 1}^{{|V_{ui}|}} {{a_{t,j'}}\mathbf{m}_{j'}} 
\end{equation}
The final hidden state $\mathbf{h}_t^{s_2}$ is the output of the second decoder GRU layer, jointly considering the word $\mathbf{s}_{t}$, the previous hidden state $\mathbf{h}_{t-1}^{s_2}$, and the attention context $\mathbf{c}_t$:
\begin{equation}
	\mathbf{h}_t^{s_2} = GRU_2(\mathbf{h}_{t-1}^{s_2}, \mathbf{s}_{t}, \mathbf{c}_t)
\end{equation}
Then we can use $\mathbf{h}_t^{s_2}$ as the input to Equation~(\ref{eq:rnn-o}) to conduct the decoding operation.

Besides using attention modeling to address and read the persona information from the the persona memory $\mathbf{M}$, we also employ the idea of Pointer Networks \cite{vinyals2015pointer} to copy the target words from the memory to form the tips. At the state $t$, we can obtain the attention weights (distribution) $\mathbf{a}_{t,:}$ on the persona memory $\mathbf{M}_{ui}$. We project $\mathbf{a}_{t,:}$ to a $|V|$-sized vector $\mathbf{\hat s}_{t+1}^{p}$ according to the word indices in $V_{ui}$. Then we design a soft gate to decide that the word $s_{t+1}$ should be generated or be copied from the memory:
\begin{equation}
	p_g = \sigma(\mathbf{v}_p^T(\mathbf{W}_{hp}^s\mathbf{h}_t^{s_2} + \mathbf{W}_{sp}^s\mathbf{s}_t + \mathbf{W}_{cp}^s\mathbf{c}_t + \mathbf{b}_{p}))
\end{equation}
where $\mathbf{v}_p \in \mathbb{R}^{d_h}$ and $p_g \in (0, 1)$. We merge the copy signal $\mathbf{\hat s}_{t+1}^{p}$ and the original output $\mathbf{\hat s}_{t+1}$ according to the gate $p_g$:
\begin{equation}
	\mathbf{\hat s}_{t+1}^\prime = p_g \times \mathbf{\hat s}_{t+1} + (1 - p_g) \times \mathbf{\hat s}_{t+1}^{p}
\end{equation}
Then the tips sampling process can be conducted on $\mathbf{\hat s}_{t+1}^\prime$.

\subsubsection{\textbf{Tips Quality Discriminator}}
Some previous works \cite{yu2017seqgan,yao2017automated} show that adversarial training strategy is beneficial to the text generation problem. To further improve the performance, we also employ this training strategy in our framework. 

The tips discriminator $D_{Tips}$ is a multilayer perceptron with the persona embeddings, the rating information, and the tips sequence as the input.
The input tips sequence can be the ground truth $S$ or the tips  $\hat S$ generated by the system. 
We propose a Bidirectional-GRU model to conduct the representation learning for $S$ and $\hat S$:
\begin{equation}
	{\mathbf{h}}^S = {{\mathord{\buildrel{\lower3pt\hbox{$\scriptscriptstyle\rightharpoonup$}} 
				\over {\mathbf{h}^S}} }}||\mathord{\buildrel{\lower3pt\hbox{$\scriptscriptstyle\leftharpoonup$}} 
		\over {\mathbf{h}^S}} 
\end{equation}
Then we combine all the information according to:
\begin{equation}
	{\mathbf{h}^q} = \tanh (\mathbf{W}_{sh}^q{\mathbf{h}^S} + \mathbf{W}_{uch}^q\mathbf{z_u^c} + \mathbf{W}_{ich}^q\mathbf{z_i^c} + \mathbf{W}_{ush}^q\mathbf{z_u^s} + \mathbf{W}_{ish}^q\mathbf{z_i^s} + \mathbf{W}_{rh}^q\mathbf{\hat r} + {\mathbf{b}^q_h})
	\nonumber
\end{equation}
Finally, we add a softmax output layer to let the model output a binary category variable:
\begin{equation}
	{\mathbf{y}^q} = \varsigma (\mathbf{W}_{hy}^q{\mathbf{h}^q} + {\mathbf{b}^q_y})
\end{equation}
We treat the ground truth $S$ as the positive instance and the sampled sequence $\hat S$ as the negative instance. So we directly let the first dimension of $\mathbf{y}^q$ represent the positive label. We define the value function as $V(S) = \mathbf{y}^q_{[0]}$.
We utilize the REINFORCE \cite{williams1992simple} method to integrate the tips quality signal $V(S)$ into the tips generation framework to conduct the parameter learning. The details can be found in the existing works \cite{yu2017seqgan,yao2017automated,li2018actor}.


\section{Experimental Setup}
\label{section4}


\subsection{Datasets}
In our experiments, we use five datasets from different domains to evaluate our framework. The ratings of all these datasets are integers in the range of $[1, 5]$.
There are four datasets from Amazon 5-core\footnote{http://jmcauley.ucsd.edu/data/amazon}: \textbf{Electronics}, \textbf{Movies \& TV}, \textbf{Clothing, Shoes and Jewelry}, and \textbf{Home and Kitchen}.
We regard the field ``summary'' as tips, and the number of tips texts is the same with the number of reviews.
Another dataset is from \textbf{Yelp} Challenge\footnote{https://www.yelp.com/dataset\_challenge}.
It is also a large-scale dataset consisting of restaurant reviews and tips.
We filter out the words with low term frequency in the tips and review texts, and build a vocabulary $\mathcal{V}$ for each dataset. We show the statistics of our datasets in Table~\ref{tbl:rec_dataset}.

\begin{table}[!t]
	\centering
	\caption[datasets]{Overview of the datasets.}
	\label{tbl:rec_dataset}
	\begin{tabular}{lrrrrr}
		\toprule
		& \textbf{Electr} &\textbf{ Movies} & \textbf{Home} & \textbf{Clothing} & \textbf{Yelp}  \\
		\midrule
		\textit{users}     & 191,522 & 123,340   & 66,212  & 39,085& 115,781 \\
		\textit{items}  & 62,333  & 49,823  & 27,991  & 22,794 & 60,224 \\
		\textit{reviews}    & 1,684,779 & 1,693,441   & 550,461  & 277,521& 1,393,257\\
		$|\mathcal{V}|$   &  37,999 & 82,805   &  23,950 &  16,297 &  82,805 \\
		\bottomrule
	\end{tabular}
\end{table}

\subsection{Evaluation Metrics}
For the evaluation of abstractive tips generation, the ground truth $s_h$ is the tips written by the user.
We use \textit{ROUGE} \cite{lin2004rouge} as our evaluation metric with standard options\footnote{ROUGE-1.5.5.pl -n 4 -w 1.2 -m -2 4 -u -c 95 -r 1000 -f A -p 0.5 -t 0}. It is a classical evaluation metric in the field of text summarization \cite{lin2004rouge}.
It counts the number of overlapping units between the generated tips and the ground truth written by users. Assuming that $s$ is the generated tips, $g_n$ is n-gram, ${C}({g_n})$ is the number of n-grams in $\tilde s$ ($s_h$ or $s$), ${C_m}({g_n})$ is the number of n-grams that appear in both $s$ and $s_h$, then the ROUGE-N score for $s$ is defined as follows:
\begin{equation}
	{ROUGE\textrm{-}N}(s) = \sum\nolimits_{{g_n} \in {s_h}} {{C_m}({g_n})/\sum\nolimits_{{g_n} \in {\tilde s}} {C({g_n})} }
\end{equation}
When $\tilde s = s_h$, we can get $ROUGE_{recall}$, and when $\tilde s = s$, we get $ROUGE_{precision}$.  We use Recall, Precision, and F-measure of ROUGE-1 (R-1), ROUGE-2 (R-2), ROUGE-L (R-L), and ROUGE-SU4 (R-SU4) to evaluate the quality of the generated tips.

For the evaluation of rating prediction,
we employ two metrics: Mean Absolute Error (\textit{MAE}) and Root
Mean Square Error (\textit{RMSE}).
Both of them are widely used for rating prediction in recommender systems. Given a predicted rating $\hat r_{u,i}$ and a ground-truth rating $r_{u,i}$ from the user $u$ for the item $i$, the RMSE is calculated as:
\begin{equation}\label{eq:expst1}
	RMSE = \sqrt {\frac{1}{N}\sum\limits_{u,i} {{{({r_{u,i}} - \hat r_{u,i})}^2}} }
\end{equation}
where $N$ indicates the number of ratings between users and items. Similarly, MAE is calculated as follows:
\begin{equation}\label{eq:expst2}
	MAE = \frac{1}{N}\sum\limits_{u,i} {\left| {{r_{u,i}} - \hat r_{u,i}} \right|}
\end{equation}

\subsection{Comparative Methods}

\textbf{Abstractive tips generation}: We compare our framework PATG with the following baseline and state-of-the-art methods:

\begin{itemize}
	\item \textbf{NRT} \cite{li2017neural}: It is a recent multi-task learning framework for rating prediction and abstractive tips generation achieving state-of-the-art performance. Latent factors for users and items are learnt during the training procedure, and are used as the context information for tips generation. NRT does not consider the persona information.
	
	\item \textbf{LexRank} \cite{erkan2004lexrank} is a classical method in the field of text summarization. Because we have obtained all the historical tips for the current user and item, then the problem can be regarded as a multi-document summarization problem. LexRank can extract a sentence as the final tips. Note that we give an advantage of this method since the ground truth ratings are used to conduct the filtering.
	
	\item \textbf{CTR$_t$}:  \textbf{C}ollaborative \textbf{T}opic \textbf{R}egression (CTR) \cite{wang2011collaborative} is proposed for rating prediction. It contains a topic model component and it can generate topics for items. Then the most topic-similar sentence from the item historical tips is extracted as the tips. 
	
	\item \textbf{HFT$_t$}: \textbf{H}idden \textbf{F}actors and Hidden \textbf{T}opics \cite{mcauley2013hidden} utilizes a topic modeling technique to model the review texts for rating prediction. Then we can design a tips extraction method HFT$_t$ using the similar technique in CTR$_t$.
\end{itemize}

\noindent\textbf{Rating prediction}: We compare of rating prediction performance with the following baseline methods:
\begin{itemize}
	
	\item \textbf{NMF}: \textbf{N}on-negative \textbf{M}atrix \textbf{F}actorization \cite{lee2001algorithms}.
	It only uses the rating matrix as the input.
	
	\item \textbf{PMF}: \textbf{P}robabilistic \textbf{M}atrix \textbf{F}actorization \cite{mnih2007probabilistic}.
	Gaussian distribution is introduced to model the latent factors for users and items.
	
	\item \textbf{LRMF}: \textbf{L}earning to \textbf{R}ank with \textbf{M}atrix \textbf{F}actorization \cite{shi2010list}. It combines a list-wise learning-to-rank algorithm with matrix factorization to improve recommendation.
	
	\item \textbf{SVD++}: It extends \textbf{S}ingular \textbf{V}alue \textbf{D}ecomposition by considering implicit feedback information for latent factor modeling \cite{koren2008factorization}.
	
	\item \textbf{URP}: \textbf{U}ser \textbf{R}ating \textbf{P}rofile modeling \cite{marlin2003modeling}. Topic models are employed to model the user preference from a generative perspective. It still only uses the rating matrix as input.
	
	\item The baseline methods used in tips quality evaluation: \textbf{NRT} \cite{li2017neural}, \textbf{CTR} \cite{wang2011collaborative}, \textbf{HFT} \cite{mcauley2013hidden}.
\end{itemize}

\noindent\textbf{Ablation experiments}: In order to demonstrate the performance of each component of our framework, we conduct the ablation experiments on the dataset Home. We compare the performance of our integrated model PATG with the models without the some designed components. We set that ``A'' denotes the aVAE model, ``M'' represents the persona memory and the Pointer Networks, and ``D'' represents the tips quality discriminator $D_{Tips}$. Then the method ``\textbf{PATG w/o A, M, D}'' means that A, M, and D are all removed and we only use the standard VAE for persona modeling. 


\subsection{Experimental Settings}
Each dataset is divided into three subsets: $80\%$, $10\%$, and $10\%$, for training, validation, and testing, respectively. All the parameters of our model  are tuned with the validation set.
After the tuning process, the number of latent factors $k$ is set to $10$ for NMF and SVD++.
The number of topics $K$ is set to $50$ for the methods using topic models.
The number of dimension for the persona embeddings is set to $100$. The dimension of the hidden size is $400$.
In our framework, the number of layers for the rating regression model is $2$, and for the tips generation model is $1$.
We set the beam size $\beta = 5$, and the maximum length $\eta = 20$.
All the neural matrix parameters in hidden layers and RNN layers are initialized from a uniform distribution between $[-0.1, 0.1]$.
We also regard the word embedding $\mathbf{E}$ used in the tips generation component as a neural parameter.
Adadelta \cite{zeiler2012adadelta} is used for gradient based optimization.

\section{Results and Discussions}
\label{section5}

\subsection{Research Questions}

The research questions in our experiments are as follows:

\begin{itemize}
	
	\item \textbf{RQ1}: What is the performance of PATG in persona-aware abstractive tips generation? (Section~\ref{sec:exp:tips})
	
	\item \textbf{RQ2}: Can the persona embeddings improve the performance of rating prediction? (Section~\ref{sec:exp:rating})
	
	\item \textbf{RQ3}: What is the performance of each component of PATG, such as VAEs, aVAE, and the persona memory? (Section~\ref{sec:exp:ablation})
	
	\item \textbf{RQ4}: Can the model generate tips that are complying with the persona information? (Section~\ref{sec:exp:persona_control}.)
	
	\item \textbf{RQ5}: Can the model generate tips that are really controlled by ratings? (Section~\ref{sec:exp:control}.)

\end{itemize}

\subsection{Abstractive Tips Generation (RQ1)}
\label{sec:exp:tips}

\begin{table*}[!htb]
	\centering
	\caption{ROUGE evaluation on the five datasets from different domains.}
	\label{tab:rouge-elects}
	\begin{tabular}{|c|c|c|c|c|c|c|c|c|c|c|c|c|c|}
		\hline
		\multirow{2}{*}{\textbf{Dataset}} &
		\multirow{2}{*}{\textbf{Method}} &
		\multicolumn{3}{c|}{\textbf{ROUGE-1}} &
		\multicolumn{3}{c|}{\textbf{ROUGE-2}} &
		\multicolumn{3}{c|}{\textbf{ROUGE-L}} &
		\multicolumn{3}{c|}{\textbf{ROUGE-SU4}} \\
		\cline{3-14}
		& & R & P & F1 & R & P & F1& R & P & F1& R & P & F1 \\
		\hline
		\multirow{5}{*}{Electronics}
		& LexRank & 10.97 & 12.93 & 11.58
		& 0.95 & 1.05 & 0.97
		& 9.96 & 11.70 & 10.50
		& 3.08 & 3.91 & 3.22 \\
		\cline{2-14}
		&  HFT$_t$ & 12.86 & 12.22 & 12.35
		& 1.10 & 1.00 & 1.03
		& 11.65 & 11.09 & 11.19
		& 3.43 & 3.10 & 3.14 \\
		\cline{2-14}
		&  CTR$_t$ & 12.69 & 11.72 & 12.02
		& 1.13 & 1.05 & 1.07
		& 11.65 & 10.74 & 11.02
		& 3.45 & 3.06 & 3.14 \\
		\cline{2-14}
		&  NRT & 12.79 & 17.55 & 13.85
		& 1.86 & 2.77 & 2.08
		& 11.80 & 15.99 & 12.70
		& 4.18 & 6.42 & 4.45 \\
		\cline{2-14}
		&  \textbf{PATG} & \textbf{13.00} & \textbf{19.26} & \textbf{14.52*}
		& \textbf{2.29} & \textbf{3.12} & \textbf{2.44*}
		& \textbf{11.91} & \textbf{17.42} & \textbf{13.24*}
		& \textbf{4.50} & \textbf{7.44} & \textbf{4.89*} \\
		\hline
		\hline
		\multirow{5}{*}{Movies\&TV}
		& LexRank & 11.10 & 13.50 & 11.89
		& 1.06 & 1.29 & 1.12
		& 10.02 & 12.12 & 10.70
		& 3.25 & 4.33 & 3.46 \\
		\cline{2-14}
		& HFT$_t$ & 11.64 & 10.26 & 11.33
		& 1.78 & 1.36 & 1.46
		& 11.42 & 8.72 & 9.67
		& 4.63 & 3.00 & 3.28 \\
		\cline{2-14}
		& CTR$_t$ & 11.37 & 10.33 & 10.68
		& 1.43 & 1.31 & 1.34
		& 10.40 & 9.44 & 9.76
		& 3.17 & 2.73 & 2.84 \\
		\cline{2-14}
		& NRT & 12.12 & 20.06 & 14.17
		& 2.29 & 3.53 & 2.55
		& 11.13 & 18.25 & 12.98
		& 4.09 & 8.15 & 4.79 \\
		\cline{2-14}
		& \textbf{PATG} & \textbf{12.46} & \textbf{21.22} & \textbf{14.63*}
		&\textbf{2.38} & \textbf{3.88} &\textbf{2.67*}
		& \textbf{11.51} & \textbf{19.25} & \textbf{14.73*}
		& \textbf{6.04} & \textbf{8.76} & \textbf{6.33*} \\
		\hline
		\hline
		\multirow{5}{*}{Home}
		& LexRank & 12.91 & 15.47 & 13.77
		& 1.73 & 2.06 & 1.82
		& 11.72 & 13.97 & 12.46
		& 3.93 & 5.02 & 4.15 \\
		\cline{2-14}
		& HFT$_t$ & 13.32 & 12.72 & 12.80
		& 1.33 & 1.23 & 1.25
		& 12.25 & 11.73 & 11.79
		& 3.63 & 3.33 & 3.34 \\
		\cline{2-14}
		& CTR$_t$ & 14.30 & 13.21 & 13.55
		& 1.73 & 1.50 & 1.58
		& 13.14 & 12.11 & 12.43
		& 4.18 & 3.66 & 3.78 \\
		\cline{2-14}
		& NRT & 11.51 & 19.91 & 13.64
		& 1.95 & 3.47 & 2.30
		& 10.64 & 18.23 & 12.57
		& 3.77 & 8.24 & 4.51 \\
		\cline{2-14}
		& \textbf{PATG} & 12.21 & \textbf{21.46} & \textbf{14.61*}
		& \textbf{2.32} & \textbf{4.32} & \textbf{2.78*}
		& 11.32 & \textbf{19.65} & \textbf{13.48*}
		& 4.03 & \textbf{8.71} & \textbf{4.82*} \\
		\hline
		\hline
		\multirow{5}{*}{Clothing}
		& LexRank & 13.31 & 12.73 & 12.85
		& 1.06 & 1.02 & 1.02
		& 11.97 & 11.43 & 11.54
		& 3.47 & 3.24 & 3.26 \\
		\cline{2-14}
		& HFT$_t$ & 13.31 & 12.73 & 12.85
		& 1.06 & 1.02 & 1.02
		& 11.97 & 11.43 & 11.54
		& 3.47 & 3.24 & 3.26 \\
		\cline{2-14}
		& CTR$_t$ & 13.79 & 13.82 & 13.37
		& 1.26 & 1.23 & 1.22
		& 12.54 & 12.14 & 12.16
		& 3.70 & 3.52 & 3.49 \\
		\cline{2-14}
		& NRT & 13.52 & 18.91 & 14.75
		& 2.11 & 2.95 & 2.31
		& 12.36 & 17.04 & 13.39
		& 4.58 & 7.04 & 4.86 \\
		\cline{2-14}
		& \textbf{PATG} & \textbf{14.45} & \textbf{21.49} & \textbf{16.14*}
		& \textbf{2.49} & \textbf{3.77} & \textbf{2.79*}
		& \textbf{13.09} & \textbf{19.24} & \textbf{14.55*}
		& \textbf{4.93} & \textbf{8.39} & \textbf{5.39*} \\
		\hline
		\hline
		\multirow{5}{*}{Yelp}
		& LexRank & 9.19 & 12.09 & 10.28
		& 1.07 & 1.33 & 1.15
		& 8.45 & 11.13 & 9.45
		& 2.65 & 3.90 & 3.01 \\
		\cline{2-14}
		& HFT$_t$ & 10.47 & 10.21 & 10.26
		& 0.91 & 0.87 & 0.88
		& 9.56 & 9.31 & 9.35
		& 2.70 & 2.57 & 2.59 \\
		\cline{2-14}
		& CTR$_t$ & 10.68 & 10.51 & 10.51
		& 0.98 & 0.94 & 0.96
		& 9.70 & 9.53 & 9.54
		& 2.77 & 2.68 & 2.68 \\
		\cline{2-14}
		& NRT & 10.98 & 17.42 & 12.71
		& 1.82 & 3.03 & 2.13
		& 9.96 & 15.76 & 11.51
		& 3.48 & 6.48 & 4.05 \\
		\cline{2-14}
		& \textbf{PATG} & \textbf{12.05} & \textbf{19.15} & \textbf{14.02*}
		& \textbf{2.15} & \textbf{3.44} & \textbf{2.47*}
		& \textbf{10.94} & \textbf{17.21} & \textbf{12.66*}
		& \textbf{3.96} & \textbf{7.15} & \textbf{4.57*} \\
		\hline
	\end{tabular}
	\\
	\vspace{1mm}
	The ``\textbf{*}'' marker denotes that PATG achieves better performance than NRT with statistical significance test with $p < 0.05$.
\end{table*}

The evaluation results of tips generation of our model and the comparative methods are given in Table~\ref{tab:rouge-elects}.
In order to capture more details, we report Recall, Precision, and F-measure (in percentage) of ROUGE-1, ROUGE-2, ROUGE-L, and ROUGE-SU4.
Our model achieves the best performance in most of the metrics among all the five datasets.
NRT does not consider persona information when generating tips. It only utilizes the learnt latent factors for users and items as the context information. Compared with NRT, our proposed framework PATG obtains better performance on all the metrics, which demonstrates that the consideration of persona information can indeed improve the tips generation performance. We also conduct statistical significance test comparing PATG and NRT and the results indicate that the improvements are significant with $p < 0.05$.

From the results, we also find that our model obtains dramatic improvements on the metric of ROUGE Precision, especially compared with the methods of LexRank, HFT$_t$, and CTR$_t$. The main reasons are that those methods are all extraction-based which just extract some original sentences from the original reviews or tips as the final tips. Therefore, the obtained tips are much longer with more noisy and redundant information. In contrast, our framework PATG as well as NRT are abstractive tips generation methods. PATG can generate more concise sentences which not only guarantee the recall metric, but also obtain better precision performance. This also fits the essential spirit of Tips.

\subsection{Rating Prediction (RQ2)}
\label{sec:exp:rating}

\begin{table*}[t]
	\begin{threeparttable}
		\centering
		\caption{ \textbf{MAE} and \textbf{RMSE} values for rating prediction. }
		\label{tab:rmse}
		\begin{tabular}{@{}lcc cc cc cc cc@{}}
			\toprule
			& \multicolumn{2}{c}{Electronics} &
			\multicolumn{2}{c}{Movies} & \multicolumn{2}{c}{Yelp}
			& \multicolumn{2}{c}{Clothing} & \multicolumn{2}{c}{Home} \\
			\cmidrule(lr){2-3}
			\cmidrule(lr){4-5}
			\cmidrule(lr){6-7}
			\cmidrule(lr){8-9}
			\cmidrule(lr){10-11}
			& MAE & RMSE  & MAE & RMSE & MAE & RMSE & MAE & RMSE  & MAE & RMSE  \\
			\midrule
			LRMF
			& 1.986\phantom{0} & 2.208\phantom{0}
			& 1.891\phantom{0} & 2.136\phantom{0}
			& 1.721\phantom{0} & 1.982\phantom{0}
			& 1.936\phantom{0} & 2.179\phantom{0}
			& 2.028\phantom{0} & 2.248\phantom{0}   \\
			PMF
			& 1.139\phantom{0} & 1.553\phantom{0}
			& 0.911\phantom{0} & 1.307\phantom{0}
			& 1.133\phantom{0} & 1.538\phantom{0}
			& 2.355\phantom{0} & 2.724\phantom{0} 
			& 1.395\phantom{0} & 1.780\phantom{0}  \\
			NMF
			& 0.869\phantom{0} & 1.266\phantom{0}
			& 0.809\phantom{0} & 1.155\phantom{0}
			& 0.961\phantom{0} & 1.136\phantom{0}
			& 0.887\phantom{0} & 1.257\phantom{0}
			& 0.830\phantom{0} & 1.220\phantom{0}  \\
			SVD++ 
			& 0.841\phantom{0} & 1.226\phantom{0}
			& 0.778\phantom{0} & 1.122\phantom{0}
			& 1.957\phantom{0} & 1.299\phantom{0}
			& 0.829\phantom{0} & 1.169\phantom{0} 
			& 0.786\phantom{0} & 1.164\phantom{0}  \\
			URP 
			& 0.875\phantom{0} & 1.185\phantom{0}
			& 0.797\phantom{0} & 1.101\phantom{0}
			& 0.973\phantom{0} & 1.246\phantom{0}
			& 0.876\phantom{0} & 1.136\phantom{0}
			& 0.831\phantom{0} & 1.135\phantom{0} \\
			CTR
			&  0.903\phantom{0} &  1.154\phantom{0}
			& 0.863\phantom{0} &  1.116\phantom{0}
			& 1.051\phantom{0} &  1.285\phantom{0}
			& 0.847\phantom{0} &  1.094\phantom{0}
			& 0.826\phantom{0} &  1.086\phantom{0} \\
			HFT 
			& 0.813\phantom{0} & 1.117\phantom{0}
			& 0.769\phantom{0} & 1.041\phantom{0}
			& 0.940\phantom{0} & 1.191\phantom{0}
			& 0.805\phantom{0} & 1.080\phantom{0}
			& 0.773\phantom{0} & 1.058\phantom{0} \\
			NRT
			& 0.823\phantom{0} & 1.108\phantom{0}
			& 0.751\phantom{0} & 1.038\phantom{0}
			& 0.935\phantom{0} & 1.187\phantom{0}
			& 0.828\phantom{0} & 1.102\phantom{0}
			& 0.779\phantom{0} & 1.058\phantom{0}  \\
			\textbf{PATG}
			& \textbf{0.747}\textbf{*} & \textbf{1.016}\textbf{*}
			& \textbf{0.740}\textbf{*} & \textbf{1.015}\textbf{*}
			& \textbf{0.866}\textbf{*} & \textbf{1.134}\textbf{*} 
			& \textbf{0.714}\textbf{*} & \textbf{0.987}\textbf{*}
			& \textbf{0.694}\textbf{*} & \textbf{0.997}\textbf{*}\\
			\bottomrule
		\end{tabular}
		\begin{tablenotes}
			\small
			\item \textbf{*} denotes that PATG achieves better performance than NRT \cite{li2017neural} with statistical significance test with $\alpha = 0.01$.
		\end{tablenotes}
	\end{threeparttable}
\end{table*}

Recall that we also design an auxiliary component, i.e. \textbf{rating prediction} to capture the sentiment information and control the sentiment of the generated tips, which is also an important aspect of the task of tips generation. Therefore we design some experiments to evaluate the performance of this component.
The rating prediction results are given in Table~\ref{tab:rmse}.
Our model consistently outperforms the best under both MAE and RMSE metrics on all datasets, thus it verifies that the generated persona embeddings are not only effective for generating better tips in the main task, but also useful for predicting accurate ratings in the auxiliary task. 
Statistical significance of differences between the performance of PATG and the recent method NRT is tested using a two-tailed paired t-test. The result shows that PATG is significantly better than NRT.

\subsection{Ablation Analysis (RQ3)}
\label{sec:exp:ablation}

\begin{table}[!t]
	\centering
	\small
	\caption{Ablation experiments on the dataset Home. R-* represents the F1-Measure of ROUGE-*.  }
	\label{tab:rouge-home-abl}
	\begin{tabular}{p{3cm} c c c c}
		\hline
		\textbf{System}  & \textbf{R-1} & \textbf{R-2} & \textbf{R-L} & \textbf{R-SU4}\\
		\hline
		PATG w/o A, M, D       & 13.76 & 2.27 & 12.64  & 4.45\\
		PATG w/o M, D & 13.99 & 2.61 & 12.95 & 4.71  \\
		PATG w/o D & 14.32 & 2.72 & 13.30 & 4.81 \\
		\textbf{PATG}  & 14.51 & 2.72 & 13.48 & 4.81 \\
		\hline
	\end{tabular}
	\vspace{-3mm}
\end{table}

Considering that we design various of components to tackle the corresponding problems of our task, and different components play different roles on our framework. In order to demonstrate the necessity and the performance of each component, we conduct the ablation experiments on the dataset ``Home''. The results are shown in Table~\ref{tab:rouge-home-abl}. Recall that ``A'' denotes the aVAE model, ``M'' represents the persona memory and the Pointer Networks, and ``D'' represents the tips quality discriminator $D_{Tips}$. It is obvious that persona modeling based on aVAE (A) can improve the tips generation performance. The persona memory and Pointer Networks (M) are very helpful to the effectiveness of our framework. The tips quality discriminator (D) can also contribute to the better performance. Among all the components, aVAE (A) as well as the persona memory and pointer network (M) contribute more to the improvements of the performance.

\subsection{Persona Controlled Generation (RQ4)}
\label{sec:exp:persona_control}
The main problem setting of this work is to generate persona-aware tips.
In order to demonstrate the quality of the generated tips,  we selected some real cases generated by our PATG from different domains for some users and items. The results are listed in Table~\ref{tbl:case}.
Although our model generates tips in an abstractive way, tips' linguistic quality is quite good. The \textbf{persona properties} of the generated tips match well with the ground truth. 
For example, in the first case, the generated tips is ``This is a great hat for the price.'', and the ground truth is ``Thanks nice quality excellent price great deal.''. Both of the sentences contain the terms ``great'' and ``price''.  In the third case, the generated tips and the ground truth  have a large overlapping with the terms ``replace my old'', and ``processor''. Interestingly, sometimes the framework can select some synonyms when conducting tips generation. For instance, the generated tips of the fourth case contains terms ``bought'' and ``for my husband''. The ground truth contains ``purchased'' and ``for a male''. 
Moreover, we also choose some generated tips with negative sentiment to conduct the sentiment correlation analysis. Take the generated tips ``Please do not buy this coffee maker.'' as an example (the last case in Table~\ref{tbl:case}), our model predicts a rating of $2.01$, which clearly shows a consistent sentiment. The ground truth tips of this example is `` They are still overpriced and all but worthless.'', which also conveys a negative sentiment. The generated tips ``The bottom line of the thin man.'' and the ground truth ``Pretty dark story in book or movie form.'' are just describing some facts, with a neutral rating $3$. Sometimes the overlapping between the generated tips and the ground truth is small, but they still convey similar information.

\begin{table}[!t]
	\centering
	\small
	\caption{Examples of the predicted ratings and the generated tips for some users and items. The first line of each group shows the generated rating and tips.
		The second line shows the ground truth.}%
	\label{tbl:case}%
	\begin{tabular}{ c | p{6.4cm}}
		\hline
		\textbf{Rating} & \ \ \ \ \ \  \  \ \ \ \ \ \ \ \ \ \ \ \ \ \ \ \ \ \ \ \ \ \ \textbf{Tips} \\

		\hline
		\textbf{\textit{5.10}} & \textbf{\textit{This is a great hat for the price.}} \\
		5 & Thanks nice quality excellent price great deal.  \\
		
		\hline
		\textbf{\textit{5.08}} & \textbf{\textit{This is a great pitcher.}} \\
		5 & Beautiful pitcher makes a great vase.  \\

		\hline
		\textbf{\textit{5.17}} & \textbf{\textit{I bought this food processor to replace my old one.}} \\
		4 & I got this about a month ago to replace my old food processor.  \\

		\hline
		\textbf{\textit{4.99}} & \textbf{\textit{These shoes are so comfortable and I bought these for my husband.}} \\
		5 & Comfortable good looking shoes purchased for a male that walks a lot.  \\
		
		\hline
		\textbf{\textit{4.81}} & \textbf{\textit{This is a great movie.}} \\
		5 & Amazing love great movie and all teen shold see it.  \\
		
		\hline
		\textbf{\textit{2.57}} &\textbf{\textit{The bottom line of the thin man.}} \\
		3 & Pretty dark story in book or movie form.  \\
		
		\hline
		\textbf{\textit{2.01}} &\textbf{\textit{Please do not buy this coffee maker.}} \\
		1 & They are still overpriced and all but worthless.  \\
		
		\hline
	\end{tabular}
	\vspace{-3mm}
\end{table}

\subsection{Rating Controlled Generation (RQ5)}
\label{sec:exp:control}

Recall that in addition to the persona embeddings as context information, rating information is also incorporated to control the sentiment of the generated tips. In order to show such ability of our framework, we design an experiment on the domain ``Home'' to demonstrate the rating controlled tips generation. Specifically, during the prediction, we manually set the rating from 1 to 5 as the sentiment context to control the generation, and meanwhile, we create a new user and a new item with $\mathbf{0}$ persona embeddings. 
This setting mimics a \textbf{cold start} case: what the tips will look like without knowing any information of users and items. 
Then the manual rating and the $\mathbf{0}$ based persona embeddings are fed into the framework to conduct tips generation. The results are shown in Table~\ref{tbl:sentiment}. Here we set the beam size to 5 so we obtain 5 decoded tips ranked by the likelihood in descending order. It is obvious that when $r > 1$, our framework can generated reasonable tips controlled by ratings. The generated tips show monotone language style, which looks odd, but in fact it is not, because we did not give any persona information of users and items as input (refer to Table \ref{tbl:case} for generated tips with rich text such as product information). 
Moreover, note that this is an artificial scenario which does not compile well with the real case. 
Table~\ref{tbl:sentiment} shows that when $r = 1$, the framework does not generate negative tips. We investigate the training dataset and find that the proportion of rating-1 records is much smaller (e.g. a quarter of rating-4 and one-thirteenth of rating-5), which may cause our model under-fitting for generating rating-1 tips.

\begin{table}[!t]
	\centering
	\caption{Rating controlled tips generation in a cold start scenario. $\hat{r}$ is the rating usd to control the sentiment. }%
	\label{tbl:sentiment}%
	\small
	\begin{tabular}{ c | p{6.8cm} }
		\hline
		\textbf{$\hat{r}$} & \textbf{Tips}\\

		\hline
		 & This is a great product. \\
		& I bought this for my mom and she loves it. \\
		\textbf{5} & I bought this for my daughter and she loves it. \\
		& I bought this for my husband and she loves it. \\
		& I bought this for my daughter. \\

		\hline
		 & This is a good product. \\
		& I bought this for my daughter for christmas.\\
		\textbf{4} & I bought this for my daughter and she loved it.\\
		& I bought this for my mom and she loved it.\\
		& I bought this for my daughter for christmas and she loves it.\\

		\hline
		& Not as good as my old one. \\
		& This is a good product.\\
	    \textbf{3} & Not as good as my old one.\\
		& I bought this for my daughter for christmas.\\
		& I bought this for my mom and she loved it.\\

		\hline
		 & Not as good as I expected. \\
		& Not as good as the original.\\
		\textbf{2} & Not as good as my old one.\\
		& I bought this for my daughter for christmas.\\
		& This is a good product.\\
		
		\hline 
    	& This is a good product.  \\
		& I bought this for my daughter for christmas. \\
		\textbf{1} & This is the third one i bought. \\
		& This is the third one i had. \\
		& I bought this for my daughter for christmas and she loves it. \\
		\hline
	\end{tabular}
	\vspace{-3mm}
\end{table}


\section{Related Work}
\label{section2}

Abstractive text generation is a challenging task.
Recently, sequence modeling based on the gated recurrent neural networks such as Long Short-Term Memory (LSTM) \cite{hochreiter1997long} and Gated Recurrent Unit (GRU) \cite{cho2014learning} demonstrates high capability in text generation related tasks, such as abstractive summarization \cite{rush2015neural,nallapati2016abstractive,li2017deep}, dialogue systems \cite{shang2015neural,cai2018skeleton} and image caption generation \cite{xu2015show}.

In the area of recommendation systems, some researchers also apply LSTM or GRU based RNN models on abstractive text generation. \citet{tang2016context} propose a framework to generate context-aware reviews. Sentiments and products are encoded into a continues semantic representation and use RNN to conduct the decoding and generation.
\citet{dong2017learning} regard users, products, and rating as attribute information and employ a attention modeling based sequence modeling framework to generate reviews.
\citet{ni2017estimating} propose to combine collaborative filtering with generative networks to jointly perform the tasks of item recommendation
and review generation. Low-dimensional user preferences and item properties are combined with a character-level LSTM model to conduct the review generation.
\citet{yao2017automated} employ the adversarial strategy to make the generated review indistinguishable from human written ones so that can improve the performance of review generation.
Although research works have been proposed for review generation, there are very few works investigating tips generation. \citet{li2017neural} propose a unified framework to jointly conduct rating prediction and abstractive tips generation. Latent factors for users and items are learnt from the multi-task learning framework and are fed into the tips generation framework as context information.

However, few works consider persona modeling and sentiment detection jointly in their frameworks. \citet{hu2017toward,liao2018quase} revised the variational auto-encoders (VAEs) based text generation model and can control the sentiment and tense of the generated reviews. But they still do not consider persona information in their model.
\citet{li2016persona} propose two methods to conduct the persona modeling for text generation in the area of dialog systems, but dialog systems have different characteristics with recommendation system. Moreover, they do not consider the sentiment information.
Different with these previous works, our proposed framework can jointly consider the persona information and the sentiment signal when conducting the abstractive tips generation. 


\section{Conclusions}
\label{section6}

We propose a framework PATG to address the problem of persona-aware tips generation.
A framework based on adversarial variational auto-encoders (aVAE) is exploited for persona modeling from the historical tips and reviews. We also design an external persona memory for directly storing the persona related words for the current user and item. 
The distilled persona embeddings are used as latent factors and are fed into the rating prediction component for detecting sentiment.
Then the persona embeddings and the sentiment information are incorporated into a recurrent neural networks (RNN) based tips generation component to control the tips generation. Experimental results show that our framework achieves better performance than the state-of-the-art models on abstractive tips generation.

\bibliographystyle{ACM-Reference-Format}
\balance 
\bibliography{www19-root}


\begin{thebibliography}{43}


\ifx \showCODEN    \undefined \def \showCODEN     #1{\unskip}     \fi
\ifx \showDOI      \undefined \def \showDOI       #1{#1}\fi
\ifx \showISBNx    \undefined \def \showISBNx     #1{\unskip}     \fi
\ifx \showISBNxiii \undefined \def \showISBNxiii  #1{\unskip}     \fi
\ifx \showISSN     \undefined \def \showISSN      #1{\unskip}     \fi
\ifx \showLCCN     \undefined \def \showLCCN      #1{\unskip}     \fi
\ifx \shownote     \undefined \def \shownote      #1{#1}          \fi
\ifx \showarticletitle \undefined \def \showarticletitle #1{#1}   \fi
\ifx \showURL      \undefined \def \showURL       {\relax}        \fi
\providecommand\bibfield[2]{#2}
\providecommand\bibinfo[2]{#2}
\providecommand\natexlab[1]{#1}
\providecommand\showeprint[2][]{arXiv:#2}

\bibitem[\protect\citeauthoryear{Bahdanau, Cho, and Bengio}{Bahdanau
  et~al\mbox{.}}{2015}]%
        {bahdanau2014neural}
\bibfield{author}{\bibinfo{person}{Dzmitry Bahdanau},
  \bibinfo{person}{Kyunghyun Cho}, {and} \bibinfo{person}{Yoshua Bengio}.}
  \bibinfo{year}{2015}\natexlab{}.
\newblock \showarticletitle{Neural machine translation by jointly learning to
  align and translate}. In \bibinfo{booktitle}{\emph{ICLR}}.
\newblock


\bibitem[\protect\citeauthoryear{Blei, Ng, and Jordan}{Blei
  et~al\mbox{.}}{2003}]%
        {blei2003latent}
\bibfield{author}{\bibinfo{person}{David~M Blei}, \bibinfo{person}{Andrew~Y
  Ng}, {and} \bibinfo{person}{Michael~I Jordan}.}
  \bibinfo{year}{2003}\natexlab{}.
\newblock \showarticletitle{Latent dirichlet allocation}.
\newblock \bibinfo{journal}{\emph{JMLR}} \bibinfo{volume}{3},
  \bibinfo{number}{Jan} (\bibinfo{year}{2003}), \bibinfo{pages}{993--1022}.
\newblock


\bibitem[\protect\citeauthoryear{Cai, Wang, Bi, Tu, Liu, Lam, and Shi}{Cai
  et~al\mbox{.}}{2018}]%
        {cai2018skeleton}
\bibfield{author}{\bibinfo{person}{Deng Cai}, \bibinfo{person}{Yan Wang},
  \bibinfo{person}{Victoria Bi}, \bibinfo{person}{Zhaopeng Tu},
  \bibinfo{person}{Xiaojiang Liu}, \bibinfo{person}{Wai Lam}, {and}
  \bibinfo{person}{Shuming Shi}.} \bibinfo{year}{2018}\natexlab{}.
\newblock \showarticletitle{Skeleton-to-Response: Dialogue Generation Guided by
  Retrieval Memory}.
\newblock \bibinfo{journal}{\emph{arXiv preprint arXiv:1809.05296}}
  (\bibinfo{year}{2018}).
\newblock


\bibitem[\protect\citeauthoryear{Card, Tan, and Smith}{Card
  et~al\mbox{.}}{2017}]%
        {card2017neural}
\bibfield{author}{\bibinfo{person}{Dallas Card}, \bibinfo{person}{Chenhao Tan},
  {and} \bibinfo{person}{Noah~A Smith}.} \bibinfo{year}{2017}\natexlab{}.
\newblock \showarticletitle{A Neural Framework for Generalized Topic Models}.
\newblock \bibinfo{journal}{\emph{arXiv preprint arXiv:1705.09296}}
  (\bibinfo{year}{2017}).
\newblock


\bibitem[\protect\citeauthoryear{Cho, Gulcehre, Bahdanau, Schwenk, and
  Bengio}{Cho et~al\mbox{.}}{2014}]%
        {cho2014learning}
\bibfield{author}{\bibinfo{person}{Kyunghyun Cho}, \bibinfo{person}{Bart van
  Merri{\"e}nboer~Caglar Gulcehre}, \bibinfo{person}{Dzmitry Bahdanau},
  \bibinfo{person}{Fethi Bougares~Holger Schwenk}, {and}
  \bibinfo{person}{Yoshua Bengio}.} \bibinfo{year}{2014}\natexlab{}.
\newblock \showarticletitle{Learning Phrase Representations using RNN
  Encoder--Decoder for Statistical Machine Translation}.
\newblock \bibinfo{journal}{\emph{EMNLP}} (\bibinfo{year}{2014}),
  \bibinfo{pages}{1724--1734}.
\newblock


\bibitem[\protect\citeauthoryear{Dong, Huang, Wei, Lapata, Zhou, and Xu}{Dong
  et~al\mbox{.}}{2017}]%
        {dong2017learning}
\bibfield{author}{\bibinfo{person}{Li Dong}, \bibinfo{person}{Shaohan Huang},
  \bibinfo{person}{Furu Wei}, \bibinfo{person}{Mirella Lapata},
  \bibinfo{person}{Ming Zhou}, {and} \bibinfo{person}{Ke Xu}.}
  \bibinfo{year}{2017}\natexlab{}.
\newblock \showarticletitle{Learning to generate product reviews from
  attributes}. In \bibinfo{booktitle}{\emph{EACL}}, Vol.~\bibinfo{volume}{1}.
  \bibinfo{pages}{623--632}.
\newblock


\bibitem[\protect\citeauthoryear{Erkan and Radev}{Erkan and Radev}{2004}]%
        {erkan2004lexrank}
\bibfield{author}{\bibinfo{person}{G{\"u}nes Erkan} {and}
  \bibinfo{person}{Dragomir~R Radev}.} \bibinfo{year}{2004}\natexlab{}.
\newblock \showarticletitle{Lexrank: Graph-based lexical centrality as salience
  in text summarization}.
\newblock \bibinfo{journal}{\emph{JAIR}}  \bibinfo{volume}{22}
  (\bibinfo{year}{2004}), \bibinfo{pages}{457--479}.
\newblock


\bibitem[\protect\citeauthoryear{Goodfellow, Pouget-Abadie, Mirza, Xu,
  Warde-Farley, Ozair, Courville, and Bengio}{Goodfellow et~al\mbox{.}}{2014}]%
        {goodfellow2014generative}
\bibfield{author}{\bibinfo{person}{Ian Goodfellow}, \bibinfo{person}{Jean
  Pouget-Abadie}, \bibinfo{person}{Mehdi Mirza}, \bibinfo{person}{Bing Xu},
  \bibinfo{person}{David Warde-Farley}, \bibinfo{person}{Sherjil Ozair},
  \bibinfo{person}{Aaron Courville}, {and} \bibinfo{person}{Yoshua Bengio}.}
  \bibinfo{year}{2014}\natexlab{}.
\newblock \showarticletitle{Generative adversarial nets}. In
  \bibinfo{booktitle}{\emph{NIPS}}. \bibinfo{pages}{2672--2680}.
\newblock


\bibitem[\protect\citeauthoryear{Goyal, Sordoni, C{\^o}t{\'e}, Ke, and
  Bengio}{Goyal et~al\mbox{.}}{2017}]%
        {goyal2017z}
\bibfield{author}{\bibinfo{person}{Anirudh Goyal Alias~Parth Goyal},
  \bibinfo{person}{Alessandro Sordoni}, \bibinfo{person}{Marc-Alexandre
  C{\^o}t{\'e}}, \bibinfo{person}{Nan Ke}, {and} \bibinfo{person}{Yoshua
  Bengio}.} \bibinfo{year}{2017}\natexlab{}.
\newblock \showarticletitle{Z-Forcing: Training Stochastic Recurrent Networks}.
  In \bibinfo{booktitle}{\emph{NIPS}}. \bibinfo{pages}{6716--6726}.
\newblock


\bibitem[\protect\citeauthoryear{Hochreiter and Schmidhuber}{Hochreiter and
  Schmidhuber}{1997}]%
        {hochreiter1997long}
\bibfield{author}{\bibinfo{person}{Sepp Hochreiter} {and}
  \bibinfo{person}{J{\"u}rgen Schmidhuber}.} \bibinfo{year}{1997}\natexlab{}.
\newblock \showarticletitle{Long short-term memory}.
\newblock \bibinfo{journal}{\emph{Neural Computation}} \bibinfo{volume}{9},
  \bibinfo{number}{8} (\bibinfo{year}{1997}), \bibinfo{pages}{1735--1780}.
\newblock


\bibitem[\protect\citeauthoryear{Hu, Yang, Liang, Salakhutdinov, and Xing}{Hu
  et~al\mbox{.}}{2017}]%
        {hu2017toward}
\bibfield{author}{\bibinfo{person}{Zhiting Hu}, \bibinfo{person}{Zichao Yang},
  \bibinfo{person}{Xiaodan Liang}, \bibinfo{person}{Ruslan Salakhutdinov},
  {and} \bibinfo{person}{Eric~P Xing}.} \bibinfo{year}{2017}\natexlab{}.
\newblock \showarticletitle{Toward controlled generation of text}. In
  \bibinfo{booktitle}{\emph{ICML}}. \bibinfo{pages}{1587--1596}.
\newblock


\bibitem[\protect\citeauthoryear{Kingma and Welling}{Kingma and
  Welling}{2014}]%
        {kingma2013auto}
\bibfield{author}{\bibinfo{person}{Diederik~P Kingma} {and}
  \bibinfo{person}{Max Welling}.} \bibinfo{year}{2014}\natexlab{}.
\newblock \showarticletitle{Auto-encoding variational bayes}. In
  \bibinfo{booktitle}{\emph{ICLR}}.
\newblock


\bibitem[\protect\citeauthoryear{Koehn}{Koehn}{2004}]%
        {koehn2004pharaoh}
\bibfield{author}{\bibinfo{person}{Philipp Koehn}.}
  \bibinfo{year}{2004}\natexlab{}.
\newblock \showarticletitle{Pharaoh: a beam search decoder for phrase-based
  statistical machine translation models}. In
  \bibinfo{booktitle}{\emph{Conference of the Association for Machine
  Translation in the Americas}}. Springer, \bibinfo{pages}{115--124}.
\newblock


\bibitem[\protect\citeauthoryear{Koren}{Koren}{2008}]%
        {koren2008factorization}
\bibfield{author}{\bibinfo{person}{Yehuda Koren}.}
  \bibinfo{year}{2008}\natexlab{}.
\newblock \showarticletitle{Factorization meets the neighborhood: a
  multifaceted collaborative filtering model}. In
  \bibinfo{booktitle}{\emph{KDD}}. ACM, \bibinfo{pages}{426--434}.
\newblock


\bibitem[\protect\citeauthoryear{Larsen, S{\o}nderby, Larochelle, and
  Winther}{Larsen et~al\mbox{.}}{2016}]%
        {larsen2016autoencoding}
\bibfield{author}{\bibinfo{person}{Anders Boesen~Lindbo Larsen},
  \bibinfo{person}{S{\o}ren~Kaae S{\o}nderby}, \bibinfo{person}{Hugo
  Larochelle}, {and} \bibinfo{person}{Ole Winther}.}
  \bibinfo{year}{2016}\natexlab{}.
\newblock \showarticletitle{Autoencoding beyond pixels using a learned
  similarity metric}. In \bibinfo{booktitle}{\emph{ICML}}.
  \bibinfo{pages}{1558--1566}.
\newblock


\bibitem[\protect\citeauthoryear{Lee and Seung}{Lee and Seung}{2001}]%
        {lee2001algorithms}
\bibfield{author}{\bibinfo{person}{Daniel~D Lee} {and}
  \bibinfo{person}{H~Sebastian Seung}.} \bibinfo{year}{2001}\natexlab{}.
\newblock \showarticletitle{Algorithms for non-negative matrix factorization}.
  In \bibinfo{booktitle}{\emph{NIPS}}. \bibinfo{pages}{556--562}.
\newblock


\bibitem[\protect\citeauthoryear{Li, Galley, Brockett, Spithourakis, Gao, and
  Dolan}{Li et~al\mbox{.}}{2016}]%
        {li2016persona}
\bibfield{author}{\bibinfo{person}{Jiwei Li}, \bibinfo{person}{Michel Galley},
  \bibinfo{person}{Chris Brockett}, \bibinfo{person}{Georgios Spithourakis},
  \bibinfo{person}{Jianfeng Gao}, {and} \bibinfo{person}{Bill Dolan}.}
  \bibinfo{year}{2016}\natexlab{}.
\newblock \showarticletitle{A Persona-Based Neural Conversation Model}. In
  \bibinfo{booktitle}{\emph{ACL}}, Vol.~\bibinfo{volume}{1}.
  \bibinfo{pages}{994--1003}.
\newblock


\bibitem[\protect\citeauthoryear{Li, Bing, and Lam}{Li et~al\mbox{.}}{2018}]%
        {li2018actor}
\bibfield{author}{\bibinfo{person}{Piji Li}, \bibinfo{person}{Lidong Bing},
  {and} \bibinfo{person}{Wai Lam}.} \bibinfo{year}{2018}\natexlab{}.
\newblock \showarticletitle{Actor-critic based training framework for
  abstractive summarization}.
\newblock \bibinfo{journal}{\emph{arXiv preprint arXiv:1803.11070}}
  (\bibinfo{year}{2018}).
\newblock


\bibitem[\protect\citeauthoryear{Li, Lam, Bing, and Wang}{Li
  et~al\mbox{.}}{2017a}]%
        {li2017deep}
\bibfield{author}{\bibinfo{person}{Piji Li}, \bibinfo{person}{Wai Lam},
  \bibinfo{person}{Lidong Bing}, {and} \bibinfo{person}{Zihao Wang}.}
  \bibinfo{year}{2017}\natexlab{a}.
\newblock \showarticletitle{Deep Recurrent Generative Decoder for Abstractive
  Text Summarization}. In \bibinfo{booktitle}{\emph{EMNLP}}.
  \bibinfo{pages}{2091--2100}.
\newblock


\bibitem[\protect\citeauthoryear{Li, Wang, Lam, Ren, and Bing}{Li
  et~al\mbox{.}}{2017b}]%
        {li2017salience}
\bibfield{author}{\bibinfo{person}{Piji Li}, \bibinfo{person}{Zihao Wang},
  \bibinfo{person}{Wai Lam}, \bibinfo{person}{Zhaochun Ren}, {and}
  \bibinfo{person}{Lidong Bing}.} \bibinfo{year}{2017}\natexlab{b}.
\newblock \showarticletitle{Salience Estimation via Variational Auto-Encoders
  for Multi-Document Summarization}. In \bibinfo{booktitle}{\emph{AAAI}}.
  \bibinfo{pages}{3497--3503}.
\newblock


\bibitem[\protect\citeauthoryear{Li, Wang, Ren, Bing, and Lam}{Li
  et~al\mbox{.}}{2017c}]%
        {li2017neural}
\bibfield{author}{\bibinfo{person}{Piji Li}, \bibinfo{person}{Zihao Wang},
  \bibinfo{person}{Zhaochun Ren}, \bibinfo{person}{Lidong Bing}, {and}
  \bibinfo{person}{Wai Lam}.} \bibinfo{year}{2017}\natexlab{c}.
\newblock \showarticletitle{Neural Rating Regression with Abstractive Tips
  Generation for Recommendation}. In \bibinfo{booktitle}{\emph{SIGIR}}. ACM,
  \bibinfo{pages}{345--354}.
\newblock


\bibitem[\protect\citeauthoryear{Li and She}{Li and She}{2017}]%
        {li2017collaborative}
\bibfield{author}{\bibinfo{person}{Xiaopeng Li} {and} \bibinfo{person}{James
  She}.} \bibinfo{year}{2017}\natexlab{}.
\newblock \showarticletitle{Collaborative variational autoencoder for
  recommender systems}. In \bibinfo{booktitle}{\emph{KDD}}. ACM,
  \bibinfo{pages}{305--314}.
\newblock


\bibitem[\protect\citeauthoryear{Liao, Bing, Li, Shi, Lam, and Zhang}{Liao
  et~al\mbox{.}}{2018}]%
        {liao2018quase}
\bibfield{author}{\bibinfo{person}{Yi Liao}, \bibinfo{person}{Lidong Bing},
  \bibinfo{person}{Piji Li}, \bibinfo{person}{Shuming Shi},
  \bibinfo{person}{Wai Lam}, {and} \bibinfo{person}{Tong Zhang}.}
  \bibinfo{year}{2018}\natexlab{}.
\newblock \showarticletitle{QuaSE: Sequence Editing under Quantifiable
  Guidance}. In \bibinfo{booktitle}{\emph{EMNLP}}. \bibinfo{pages}{3855--3864}.
\newblock


\bibitem[\protect\citeauthoryear{Lin}{Lin}{2004}]%
        {lin2004rouge}
\bibfield{author}{\bibinfo{person}{Chin-Yew Lin}.}
  \bibinfo{year}{2004}\natexlab{}.
\newblock \showarticletitle{ROUGE: A Package for Automatic Evaluation of
  Summaries}. In \bibinfo{booktitle}{\emph{Text Summarization Branches Out-ACL
  Workshop}}. \bibinfo{pages}{74--81}.
\newblock


\bibitem[\protect\citeauthoryear{Marlin}{Marlin}{2003}]%
        {marlin2003modeling}
\bibfield{author}{\bibinfo{person}{Benjamin~M Marlin}.}
  \bibinfo{year}{2003}\natexlab{}.
\newblock \showarticletitle{Modeling user rating profiles for collaborative
  filtering}. In \bibinfo{booktitle}{\emph{NIPS}}. \bibinfo{pages}{627--634}.
\newblock


\bibitem[\protect\citeauthoryear{McAuley and Leskovec}{McAuley and
  Leskovec}{2013}]%
        {mcauley2013hidden}
\bibfield{author}{\bibinfo{person}{Julian McAuley} {and} \bibinfo{person}{Jure
  Leskovec}.} \bibinfo{year}{2013}\natexlab{}.
\newblock \showarticletitle{Hidden factors and hidden topics: understanding
  rating dimensions with review text}. In \bibinfo{booktitle}{\emph{RecSys}}.
  ACM, \bibinfo{pages}{165--172}.
\newblock


\bibitem[\protect\citeauthoryear{Mescheder, Nowozin, and Geiger}{Mescheder
  et~al\mbox{.}}{2017}]%
        {mescheder2017adversarial}
\bibfield{author}{\bibinfo{person}{Lars Mescheder}, \bibinfo{person}{Sebastian
  Nowozin}, {and} \bibinfo{person}{Andreas Geiger}.}
  \bibinfo{year}{2017}\natexlab{}.
\newblock \showarticletitle{Adversarial Variational Bayes: Unifying Variational
  Autoencoders and Generative Adversarial Networks}. In
  \bibinfo{booktitle}{\emph{ICML}}. \bibinfo{pages}{2391--2400}.
\newblock


\bibitem[\protect\citeauthoryear{Nallapati, Zhou, dos Santos, Gulcehre, and
  Xiang}{Nallapati et~al\mbox{.}}{2016}]%
        {nallapati2016abstractive}
\bibfield{author}{\bibinfo{person}{Ramesh Nallapati}, \bibinfo{person}{Bowen
  Zhou}, \bibinfo{person}{Cicero dos Santos}, \bibinfo{person}{Caglar
  Gulcehre}, {and} \bibinfo{person}{Bing Xiang}.}
  \bibinfo{year}{2016}\natexlab{}.
\newblock \showarticletitle{Abstractive Text Summarization using
  Sequence-to-sequence RNNs and Beyond}. In
  \bibinfo{booktitle}{\emph{Proceedings of The 20th SIGNLL Conference on
  Computational Natural Language Learning}}. \bibinfo{pages}{280--290}.
\newblock


\bibitem[\protect\citeauthoryear{Ni, Lipton, Vikram, and McAuley}{Ni
  et~al\mbox{.}}{2017}]%
        {ni2017estimating}
\bibfield{author}{\bibinfo{person}{Jianmo Ni}, \bibinfo{person}{Zachary~C
  Lipton}, \bibinfo{person}{Sharad Vikram}, {and} \bibinfo{person}{Julian
  McAuley}.} \bibinfo{year}{2017}\natexlab{}.
\newblock \showarticletitle{Estimating Reactions and Recommending Products with
  Generative Models of Reviews}. In \bibinfo{booktitle}{\emph{IJCNLP}},
  Vol.~\bibinfo{volume}{1}. \bibinfo{pages}{783--791}.
\newblock


\bibitem[\protect\citeauthoryear{Ren, Liang, Li, Wang, and de~Rijke}{Ren
  et~al\mbox{.}}{2017}]%
        {rensocial2017}
\bibfield{author}{\bibinfo{person}{Zhaochun Ren}, \bibinfo{person}{Shangsong
  Liang}, \bibinfo{person}{Piji Li}, \bibinfo{person}{Shuaiqiang Wang}, {and}
  \bibinfo{person}{Maarten de Rijke}.} \bibinfo{year}{2017}\natexlab{}.
\newblock \showarticletitle{Social collaborative viewpoint regression with
  explainable recommendations}. In \bibinfo{booktitle}{\emph{WSDM}}. ACM,
  \bibinfo{pages}{485--494}.
\newblock


\bibitem[\protect\citeauthoryear{Rush, Chopra, and Weston}{Rush
  et~al\mbox{.}}{2015}]%
        {rush2015neural}
\bibfield{author}{\bibinfo{person}{Alexander~M Rush}, \bibinfo{person}{Sumit
  Chopra}, {and} \bibinfo{person}{Jason Weston}.}
  \bibinfo{year}{2015}\natexlab{}.
\newblock \showarticletitle{A Neural Attention Model for Abstractive Sentence
  Summarization}. In \bibinfo{booktitle}{\emph{EMNLP}}.
  \bibinfo{pages}{379--389}.
\newblock


\bibitem[\protect\citeauthoryear{Salakhutdinov and Mnih}{Salakhutdinov and
  Mnih}{2007}]%
        {mnih2007probabilistic}
\bibfield{author}{\bibinfo{person}{Ruslan Salakhutdinov} {and}
  \bibinfo{person}{Andriy Mnih}.} \bibinfo{year}{2007}\natexlab{}.
\newblock \showarticletitle{Probabilistic Matrix Factorization.}. In
  \bibinfo{booktitle}{\emph{NIPS}}. \bibinfo{pages}{1--8}.
\newblock


\bibitem[\protect\citeauthoryear{Shang, Lu, and Li}{Shang
  et~al\mbox{.}}{2015}]%
        {shang2015neural}
\bibfield{author}{\bibinfo{person}{Lifeng Shang}, \bibinfo{person}{Zhengdong
  Lu}, {and} \bibinfo{person}{Hang Li}.} \bibinfo{year}{2015}\natexlab{}.
\newblock \showarticletitle{Neural Responding Machine for Short-Text
  Conversation}. In \bibinfo{booktitle}{\emph{ACL}}, Vol.~\bibinfo{volume}{1}.
  \bibinfo{pages}{1577--1586}.
\newblock


\bibitem[\protect\citeauthoryear{Shi, Larson, and Hanjalic}{Shi
  et~al\mbox{.}}{2010}]%
        {shi2010list}
\bibfield{author}{\bibinfo{person}{Yue Shi}, \bibinfo{person}{Martha Larson},
  {and} \bibinfo{person}{Alan Hanjalic}.} \bibinfo{year}{2010}\natexlab{}.
\newblock \showarticletitle{List-wise learning to rank with matrix
  factorization for collaborative filtering}. In
  \bibinfo{booktitle}{\emph{RecSys}}. \bibinfo{pages}{269--272}.
\newblock


\bibitem[\protect\citeauthoryear{Tang, Yang, Carton, Zhang, and Mei}{Tang
  et~al\mbox{.}}{2016}]%
        {tang2016context}
\bibfield{author}{\bibinfo{person}{Jian Tang}, \bibinfo{person}{Yifan Yang},
  \bibinfo{person}{Sam Carton}, \bibinfo{person}{Ming Zhang}, {and}
  \bibinfo{person}{Qiaozhu Mei}.} \bibinfo{year}{2016}\natexlab{}.
\newblock \showarticletitle{Context-aware Natural Language Generation with
  Recurrent Neural Networks}.
\newblock \bibinfo{journal}{\emph{arXiv preprint arXiv:1611.09900}}
  (\bibinfo{year}{2016}).
\newblock


\bibitem[\protect\citeauthoryear{Vinyals, Fortunato, and Jaitly}{Vinyals
  et~al\mbox{.}}{2015}]%
        {vinyals2015pointer}
\bibfield{author}{\bibinfo{person}{Oriol Vinyals}, \bibinfo{person}{Meire
  Fortunato}, {and} \bibinfo{person}{Navdeep Jaitly}.}
  \bibinfo{year}{2015}\natexlab{}.
\newblock \showarticletitle{Pointer networks}. In
  \bibinfo{booktitle}{\emph{NIPS}}. \bibinfo{pages}{2692--2700}.
\newblock


\bibitem[\protect\citeauthoryear{Wang and Blei}{Wang and Blei}{2011}]%
        {wang2011collaborative}
\bibfield{author}{\bibinfo{person}{Chong Wang} {and} \bibinfo{person}{David~M
  Blei}.} \bibinfo{year}{2011}\natexlab{}.
\newblock \showarticletitle{Collaborative topic modeling for recommending
  scientific articles}. In \bibinfo{booktitle}{\emph{KDD}}. ACM,
  \bibinfo{pages}{448--456}.
\newblock


\bibitem[\protect\citeauthoryear{Williams}{Williams}{1992}]%
        {williams1992simple}
\bibfield{author}{\bibinfo{person}{Ronald~J Williams}.}
  \bibinfo{year}{1992}\natexlab{}.
\newblock \showarticletitle{Simple statistical gradient-following algorithms
  for connectionist reinforcement learning}.
\newblock \bibinfo{journal}{\emph{Machine Learning}} \bibinfo{volume}{8},
  \bibinfo{number}{3-4} (\bibinfo{year}{1992}), \bibinfo{pages}{229--256}.
\newblock


\bibitem[\protect\citeauthoryear{Xu, Ba, Kiros, Cho, Courville, Salakhudinov,
  Zemel, and Bengio}{Xu et~al\mbox{.}}{2015}]%
        {xu2015show}
\bibfield{author}{\bibinfo{person}{Kelvin Xu}, \bibinfo{person}{Jimmy Ba},
  \bibinfo{person}{Ryan Kiros}, \bibinfo{person}{Kyunghyun Cho},
  \bibinfo{person}{Aaron Courville}, \bibinfo{person}{Ruslan Salakhudinov},
  \bibinfo{person}{Rich Zemel}, {and} \bibinfo{person}{Yoshua Bengio}.}
  \bibinfo{year}{2015}\natexlab{}.
\newblock \showarticletitle{Show, attend and tell: Neural image caption
  generation with visual attention}. In \bibinfo{booktitle}{\emph{ICML}}.
  \bibinfo{pages}{2048--2057}.
\newblock


\bibitem[\protect\citeauthoryear{Yao, Viswanath, Cryan, Zheng, and Zhao}{Yao
  et~al\mbox{.}}{2017}]%
        {yao2017automated}
\bibfield{author}{\bibinfo{person}{Yuanshun Yao}, \bibinfo{person}{Bimal
  Viswanath}, \bibinfo{person}{Jenna Cryan}, \bibinfo{person}{Haitao Zheng},
  {and} \bibinfo{person}{Ben~Y Zhao}.} \bibinfo{year}{2017}\natexlab{}.
\newblock \showarticletitle{Automated Crowdturfing Attacks and Defenses in
  Online Review Systems}. In \bibinfo{booktitle}{\emph{CCS}}. ACM,
  \bibinfo{pages}{1143--1158}.
\newblock


\bibitem[\protect\citeauthoryear{Yu, Zhang, Wang, and Yu}{Yu
  et~al\mbox{.}}{2017}]%
        {yu2017seqgan}
\bibfield{author}{\bibinfo{person}{Lantao Yu}, \bibinfo{person}{Weinan Zhang},
  \bibinfo{person}{Jun Wang}, {and} \bibinfo{person}{Yong Yu}.}
  \bibinfo{year}{2017}\natexlab{}.
\newblock \showarticletitle{SeqGAN: Sequence Generative Adversarial Nets with
  Policy Gradient.}. In \bibinfo{booktitle}{\emph{AAAI}}.
  \bibinfo{pages}{2852--2858}.
\newblock


\bibitem[\protect\citeauthoryear{Zeiler}{Zeiler}{2012}]%
        {zeiler2012adadelta}
\bibfield{author}{\bibinfo{person}{Matthew~D Zeiler}.}
  \bibinfo{year}{2012}\natexlab{}.
\newblock \showarticletitle{ADADELTA: an adaptive learning rate method}.
\newblock \bibinfo{journal}{\emph{arXiv preprint arXiv:1212.5701}}
  (\bibinfo{year}{2012}).
\newblock


\bibitem[\protect\citeauthoryear{Zhao, Zhao, and Eskenazi}{Zhao
  et~al\mbox{.}}{2017}]%
        {zhao2017learning}
\bibfield{author}{\bibinfo{person}{Tiancheng Zhao}, \bibinfo{person}{Ran Zhao},
  {and} \bibinfo{person}{Maxine Eskenazi}.} \bibinfo{year}{2017}\natexlab{}.
\newblock \showarticletitle{Learning Discourse-level Diversity for Neural
  Dialog Models using Conditional Variational Autoencoders}. In
  \bibinfo{booktitle}{\emph{ACL}}. \bibinfo{pages}{654--664}.
\newblock


\end{thebibliography}

\end{document}